\documentclass{article}
\usepackage{arxiv}
\usepackage{hyperref}
\usepackage{graphicx}
\usepackage{xcolor}
\usepackage{algorithm}
\usepackage[noend]{algorithmic}
\usepackage{pdflscape}
\usepackage{amsmath,amsthm,amsfonts}

\usepackage{natbib}

\AtBeginDocument{%
  }

\newcommand{\aneta}[1]{}
\newcommand{\frank}[1]{}
\newcommand{\kokila}[1]{}
\newcommand{\ignore}[1]{}

\begin{document}
\title{Trust Region-Based Bayesian Optimisation to Discover Diverse Solutions}

%Optimisation and Logistics, School of Computer and Mathematical Sciences, The University of Adelaide
\author{
        {Kokila Perera} \\
        Optimisation and Logistics\\
	School of Computer and Mathematical Sciences\\
        The University of Adelaide\\
        Adelaide, Australia \\
        \And
	{Frank Neumann} \\
        Optimisation and Logistics\\
	School of Computer and Mathematical Sciences\\
        The University of Adelaide\\
        Adelaide, Australia \\
        \And
	{Aneta Neumann} \\
        Optimisation and Logistics\\
	School of Computer and Mathematical Sciences\\
        The University of Adelaide\\
        Adelaide, Australia \\
}
\maketitle

\begin{abstract} 
%\aneta{It would be great if you could please prepare a poster for FOGA (similar as for PPSN)} Poster: Noted.
%\aneta{You can also use the following short form for GECCO, PPSN: Proc. of GECCO 20xx, in the References in case space is limited.} Noted.
Bayesian optimisation (BO) is a surrogate-based optimisation technique that efficiently solves expensive black-box functions with small evaluation budgets. Recent studies consider trust regions to improve the scalability of BO approaches when the problem space scales to more dimensions. Motivated by this research, we explore the effectiveness of trust region-based BO algorithms for diversity optimisation in different dimensional black box problems. 
We propose diversity optimisation approaches extending TuRBO1, which is the first BO method that uses a trust region-based approach for scalability. We extend TuRBO1 as divTuRBO1, which finds an optimal solution while maintaining a given distance threshold relative to a reference solution set. We propose two approaches to find diverse solutions for black-box functions by combining divTuRBO1 runs in a sequential and an interleaving fashion. We conduct experimental investigations on the proposed algorithms and compare their performance with that of the baseline method, ROBOT (rank-ordered Bayesian optimisation with trust regions). We evaluate proposed algorithms on benchmark functions with dimensions 2 to 20. Experimental investigations demonstrate that the proposed methods perform well, particularly in larger dimensions, even with a limited evaluation budget.  
\end{abstract}


\section{Introduction}

Discovering a diverse set of high-quality solutions is beneficial in many real-world optimisation problems. In practical applications, unforeseen changes to the problem conditions can occur, such as changes in available resources. If the problem changes its previously known circumstances after optimisation, the outcome solutions can become useless. Therefore, we have to repeat the optimisation considering the new information. However, if one has access to a structurally diverse set of high-performing solutions instead of a sole optimal solution, it would be easier to incorporate the new information. Decision-makers can make comparisons and inspect the solutions under additional criteria. Considering diversity in optimisation has been explored in evolutionary computation for a while~\cite{10.1145/3512290.3528851,10.1145/3512290.3528862,10.1007/978-3-031-14721-0_29}. Diversity optimisation has many practical applications in various domains, including communication networks~\cite{10.1145/3583131.3590405,10.1145/3638529.3654184}, healthcare~\cite{10.1007/978-3-031-14714-2_18,DBLP:conf/gecco/NikfarjamSND024}, robot learning \cite{dynamicRobot10.1145/3583133.3590625}, game design~\cite{games10.1007/s10107-024-02126-8}, computational creativity~\cite{DBLP:conf/gecco/NeumannGDN018,DBLP:conf/gecco/NeumannG0019} and materials science~\cite{crystalStructure10.1145/3638529.3654048}.

Diversity is a new topic to Bayesian optimisation (BO) compared to evolutionary computation \cite{ROBOT-pmlr-v206-maus23a}. BO is popular because of its data-efficient performance. BO methods are generally designed for single-objective functions and do not focus on diversity. These methods are based on a surrogate model and do not require the full knowledge of the functional features of the target function \cite{recent-advances-10.1145/3582078}. BO is ideal for black-box functions that are expensive to evaluate and predominantly used for machine learning hyperparameter tuning, bioinformatics, chemical engineering, and many other applications \cite{10.3389/fpls.2022.1050198,NIPS2012_05311655,pmlr-v162-stanton22a,WANG2022100728}. BO is an active research field, and recent developments in the field address the limitations of the classical BO methods. For example, recently proposed BO methods in \citet{TuRBO-10.5555/3454287.3454780} and \citet{MORBO-pmlr-v180-daulton22a} cater for multi-objective functions and higher-dimensional search spaces \cite{TuRBO-10.5555/3454287.3454780,MORBO-pmlr-v180-daulton22a}.

%%%%%%% Literature on BO for diversity
In recent BO literature, a few studies have considered diversity optimisation \cite{kent2023bopelites,kentInteractive2023, ROBOT-pmlr-v206-maus23a}. BOP-Elites is the first BO algorithm that caters for diversity optimisation \cite{kent2023bopelites}. This algorithm aims to discover diverse elite solutions for target problems. BOP-Elites uses a surrogate model that captures both the objective and behavioural descriptors. These behavioural descriptors can be used after convergence to predict solutions for any behaviour. However, BOP-Elites do not perform well when addressing problems with dimensions $> 10$ \cite{kent2023bopelites}. Later, Kent et al. (2023b) \cite{kentInteractive2023} present an interactive Bayesian optimisation approach to diversity, where decision-makers are involved during optimisation to specify the region of interest to explore and target specific behaviours. This method has also been experimented with only up to 10 dimensions, and the success of the algorithms heavily depends on the availability of human expertise. 

The most recent BO algorithm with the diversity optimisation feature is ROBOT \cite{ROBOT-pmlr-v206-maus23a}. ROBOT makes use of trust region-based optimisation, which is a recently emerged concept for improving the scalability of BO methods \cite{TuRBO-10.5555/3454287.3454780}. The trust regions are used in these BO methods to utilise local optimisation, exploiting the promising areas in the search space. They help to reduce the over-exploration, a common problem when using BO for large-scale problems. Thus, trust regions help balance exploration and exploitation in BO. In the ROBOT algorithm, multiple trust regions are used simultaneously to find high-quality solutions that are diversely located over the search space. The number of trust regions required by this algorithm is equal to the desired number of diverse solutions. When this number is large, ROBOT can be computationally expensive and may not perform well with a small budget.

\subsection{Our Contributions}
%This paper explores BO approaches that efficiently produce a set of structurally diverse, high-quality solutions.
This work is motivated by the scalable BO approach TuRBO~\cite{TuRBO-10.5555/3454287.3454780}, which, in particular, with a single trust region (TuRBO1), generates successful results with a reasonable evaluation budget and CPU time \cite{santoni2023comparison}. Considering these properties of TuRBO1, we adapt it for diversity optimisation to address optimisation problems efficiently. 

First, we propose an extended algorithm, called divTuRBO1, which optimises the problem while maintaining diversity with a given input set. This input set of individuals restricts the optimisation process of divTuRBO1. This set of individuals can be utilised as a mediator between multiple runs of divTuRBO1, which together discover diverse solutions for an optimisation problem. We propose two approaches to combine divTuRBO runs: a sequential and an interleaving approach. Respectively, we introduce two algorithms, divTuRBO1-seq and divTuRBO1-int, which utilise divTuRBO1 runs to address diversity optimisation. 

We conduct experimental evaluations on the proposed diversity approaches, using ROBOT \cite{ROBOT-pmlr-v206-maus23a} as the baseline method for comparisons. For experimental investigations, we compare the performance of algorithms on benchmark black-box optimisation functions with dimensions from 2 to 20, considering different diversity threshold levels. The detailed experimental analysis compares the performance of the proposed methods and ROBOT over these benchmark functions. Moreover, we consider the influence of the evaluation budget on these algorithms as well as the number of phases used in the interleaving approach when generating diverse results.

The remaining content of this paper is organised as follows. Section \ref{sec:prelim} introduces the core concepts related to this work, including preliminaries on BO, trust region-based optimisation, and diversity. We explain how we derive the new algorithmic approaches and describe them in Section \ref{sec:algo}. In Section \ref{sec:experiments}, we describe the design of experimental settings used to evaluate the algorithms. Finally, we discuss the results of our experimental evaluation in detail in Section \ref{sec:results}, followed by the concluding remarks in Section \ref{sec:conclusion}.
\section{Background and Preliminary Concepts}\label{sec:prelim} % prelims and background
In this study, we propose diversity optimisation methods that incorporate diversity principles into TuRBO1. TuRBO1 is an efficient BO method that utilises local optimisation via trust regions to find solutions in higher-dimensional optimisation problems efficiently. In this section, we briefly introduce the underlying concepts of Bayesian optimisation and diversity optimisation. 

\subsection{Bayesian Optimisation}
Bayesian Optimisation (BO) methods are sequential optimisation methods with two main components: a probabilistic surrogate model and an acquisition function. The surrogate model is usually a Gaussian process model that predicts the promising search points. This model reduces the evaluations of expensive black-box functions. One or more of the predicted search points are selected by the acquisition function. The role of the acquisition function is to trade off exploration and exploitation during optimisation. During optimisation, the surrogate model is trained based on the black-box function evaluations of the new data points selected in each iteration. However, in search spaces with 20 or more dimensions, surrogate model predictions may lead to an overemphasis on exploration, and the algorithm may fail to find high-quality solutions within the given budget. 

The trust region-based BO methods were introduced to address the above problem.
A trust region is a hypercube that marks the promising sub-space in the global search space. The algorithm restricts the exploration of new search points to this region as the surrogate predictions adopt the value intervals (for each dimension) covered by this hypercube. The trust region employs an adaptive approach, positioning itself in better locations as new solutions are discovered. The lengths of the trust region self-adjust depending on the success of the predictions made in each iteration. These adjustments guide the surrogate model in making localised predictions and increase the scalability of the algorithms to perform well as the number of dimensions in the problem increases. The algorithms can use a single or multiple trust regions collaboratively to find optimal solutions. For example, TuRBO has different variations as TuRBO1 and TuRBO-M, which use a single or multiple ($M>1$) trust regions for optimisation, respectively.

\ignore{
\begin{algorithm}
    \begin{algorithmic}
    \caption{TuRBO1\label{alg:turbo}}
    \REQUIRE $n_{0}$ (initial batch size) and $n$ (batch size)
    \STATE $X_{global} \gets \emptyset$
    \WHILE{$n_{eval} \leq max_{eval}$}
        \STATE initialise or reset TR lengths
        \STATE $X_{tr} \gets $ randomly generated $n_{0}$ solution points
        \STATE $X_{global} \gets X_{global} \cup X_{tr}$
        \WHILE{$n_{eval} \leq max_{eval}$  $\AND $ the TR is sufficiently large}
            \STATE $x_{center} \gets $ The best point in $X_{tr}$
            \STATE Pass the TR parameters ($x_{center}$ and lengths) to the surrogate model
            \STATE $X_{next} \gets$ the best $n$ points predicted by the surrogate model
            \STATE Adjust the TR lengths according to the success/failure of $X_{next}$
            \STATE Add $X_{next}$ to $X_{global}$ and $X_{tr}$
            \STATE Update $X_{tr}$, $X_{global}$ and the surrogate model with new points $X_{next}$
        \ENDWHILE
    \ENDWHILE
    \RETURN The best solution in $X_{global}$
\end{algorithmic}
\end{algorithm}
}
TuRBO1 \cite{TuRBO-10.5555/3454287.3454780} starts with a set of random solutions from the global search space. Initially, the center of the trust region is placed at the best solution among the initial solutions, and its lengths are set according to the algorithm parameters. In each iteration, the trust region is moved by centering it on the best solution found within that trust region, and lengths are adjusted to reflect the success or failure of the predictions. It is considered a success if the trust region provides better predictions than the previous solutions. Otherwise, it is considered a failure. In a successful iteration, the hypercube lengths of the trust region are increased. 
In contrast, the trust region lengths are reduced in failed iterations. If the trust region fails repeatedly and its length reaches the minimum threshold value, the trust region is abandoned and restarted with random solutions from the global search space. This allows the optimisation to escape the local search if it gets trapped in an unfavourable region. The dynamic trust region guides the surrogate model to focus on the promising areas of the solution space as new predictions are made. Trust regions enable BO methods to strike a balance between exploration and exploitation, staying close to the actual objective while improving efficiency and maintaining accuracy.

\subsection{Diverse optimisation for black-box functions}
\label{sec:diversity}

The diversity optimisation goals can be defined in different ways. One approach is to define a minimum threshold for the quality of the solutions, identify a certain number of solutions from the search space that meet this quality threshold, and maintain the maximum possible distance between each other. Another approach is to define a minimum threshold for diversity and optimise the problem to identify a certain number of best solutions that maintain a minimum distance from each other as enforced by the threshold value. In line with  \citet{ROBOT-pmlr-v206-maus23a}, we consider the second approach to define diversity optimisation goals. %Indicates that we consider only fixed tau values. 

Given $f$ as a target function defined on a search space $S$ with $D$ dimensions, it takes the form $f \colon S \subseteq \mathbb{R}^D \rightarrow \mathbb{R}$. We can define the distance between two solutions $x,y\in S$ using the Euclidean distance measure as follows,
$$div(x,y)=||x - y|| = \sqrt{(x- y)^2}.$$
%The goal of diversity optimisation of these problems is to find $m$ best solutions for the problem, which maintains a minimum distance threshold of value $\tau$ with each other. We use the Euclidean distance measure to define the distance between the solution points of a particular BBOB function. Given two candidate points $x$ and $y$ of a given function in $D$ dimensional space, we calculate the distance between $x$ and $y$ as follows:
Consider a scenario where we need to compare two sets of solutions against each other. We can introduce a minimum distance threshold, denoted as $\tau$, to define a desired level of diversity between the two sets. Assume that solutions in one of the sets $X_{div}$ already maintain the minimum distance between each other as $\tau$. Given the other set of solution vectors as $X$, and we can define the subset $X_{\tau} \subseteq X$ that satisfy the minimum distance threshold $\tau$ with $X_{div}$ as follows,
$$X_{\tau} = \{x|div(x,y)\geq\tau, x \in X, y \in X_{div}\}.$$

Furthermore, we can identify the best solution in $X$ that maintains the diversity threshold $\tau$ with $X_{div}$. For a given black-box function $f$, the distance threshold $\tau$ and a set of diverse elites for $f$ as $X_{div}$, the best diverse solution $x_{best}$ in $X$ can be defined as follows, 
\begin{equation}
    \label{eq:best-div-solution}
    x_{best} = \left\{
    \begin{array}{lcl}
        \arg_x \min_{x \in X_{\tau}} f(x) && X_{\tau}\neq \emptyset
        \\
        \arg_x \max_{x \in X,y \in X_{div}} div(x,y)
        && X_{\tau} = \emptyset
    \end{array} \right.
\end{equation}

According to this definition, if none of the solutions in $X$ satisfy the minimum distance criterion, the solution that is farthest away from the solutions in $X_{div}$ is selected as the best diverse solution despite its value.

\begin{algorithm}[!t]
    \begin{algorithmic}[1]
    \caption{divTuRBO1}\label{alg:divTuRBO1}
    \REQUIRE $n_{0}$ (initial batch size), $n$ (batch size) and $X_{div}$ (set of diverse elites)
    \STATE $X_{global} \gets \emptyset$
    \WHILE{remaining evaluation budget $> 0$}
        \STATE initialise or reset the TR lengths
        \STATE $X_{tr} \gets $ randomly generated $n_{0}$ solution points
        \STATE $X_{global} \gets X_{global} \cup X_{tr}$
        \WHILE{remaining evaluation budget $> 0$  $\AND $ TR is sufficiently large}
            \STATE $x_{center} \gets $ The best $x \in X_{tr}$ that is also diverse from $x' \in X_{div}$
            \STATE Pass the TR parameters ($x_{center}$ and lengths) to the surrogate model
            \STATE $X_{next} \gets$ best $n$ points predicted by the surrogate model that maintains the diversity with $X_{div}$
            \STATE Increase/decrease TR lengths according to the success/failure of $X_{next}$
            \STATE Update $X_{tr}$, $X_{global}$ and the surrogate model with new points $X_{next}$
        \ENDWHILE
    \ENDWHILE
    \RETURN The best diverse solution in  $X_{global}$
\end{algorithmic}
\end{algorithm}

\begin{algorithm}[!t]
    \begin{algorithmic}[1]
    \caption{divTuRBO1-seq}\label{alg:divTuRBO1-seq}
    \REQUIRE $m$ (number of diverse solutions), $B$ (total budget)
    \STATE $X_{div} \gets \emptyset$
    \STATE $B' \gets B/m$ 
    \FOR{$i\gets 1$ \textbf{to} $m$}
        \STATE Run divTuRBO1 with $X_{div}$ with $B'$ evaluations
        \STATE $x \gets $ the best diverse solution from the run
        \STATE $X_{div} \cup x $
    \ENDFOR
    \RETURN $X_{div}$
\end{algorithmic}
\end{algorithm}

\begin{algorithm}[!t]
    \begin{algorithmic}[1]
    \caption{divTuRBO1-int}\label{alg:divTuRBO1-int}
    \REQUIRE $m$ (number of diverse solutions), $B$ (total budget), $max_{phases}$ (number of interleaving phases per run)
    \STATE $X_{div} \gets \emptyset$
    \STATE $B' \gets B/(m\cdot max_{phases})$ 
    \FOR{$phase \gets 1$ \textbf{to} $max_{phases}$}
        \FOR{$i\gets 1$ \textbf{to} $m$}
            \STATE Start/resume $i^{th}$ run of divTuRBO1 with budget $B'$
            \STATE $x \gets $ the best diverse solution from the run
            \IF{$phase=1$}    
                \STATE $X_{div} \cup x $
            \ELSE
                \STATE $X_{div}[i] \gets x$
            \ENDIF
        \ENDFOR
    \ENDFOR
    \RETURN $X_{div}$
\end{algorithmic}
\end{algorithm}

\section{Algorithms}\label{sec:algo}

In this section, we introduce the algorithms proposed in this paper. We propose new diversity optimisation methods derived from TuRBO \cite{TuRBO-10.5555/3454287.3454780} with a single trust region (denoted as TuRBO1). This base algorithm can be considered as a simple form of the BO algorithms that use trust regions. Since there is only one trust region, computations are less expensive than those that use multiple trust regions simultaneously. As Santoni et al. \cite{santoni2023comparison} show, TuRBO1 performs quite well for BBOB functions with small evaluation budgets. For the specific steps in TuRBO1, we refer the reader to \cite{TuRBO-10.5555/3454287.3454780}.

%divTuRBO1: Extending TuRBO1 to address diversity constraints
\subsection{Implementing diversity with trust region-based BO}
In this work, we derive a diversity optimisation approach based on TuRBO1, reffered to as divTuRBO1. The new algorithm requires an input set of diverse elites and a minimum distance threshold. divTuRBO1 focuses on finding one optimal solution that maintains the diversity criterion against the input set. The extended algorithm addresses the diversity criterion at two key stages of the base algorithm, TuRBO1. These two stages are the selection of the center point of the trust region and the selection of candidate points. 

%This placement of the trust region guides the prediction of the next candidates by the surrogate model. 
Firstly, we change the criterion for selection of the center point of the trust region. In TuRBO1, the single trust region is updated in each iteration to be centered on the solution with the best objective value among the solutions in the iteration. Thus, TuRBO1 predicts new solution points close to this best solution chosen as the center. In the extended algorithm divTuRBO1, the new candidate points should also maintain the distance from the input set of diverse elites. Thus, we alter the process of selecting the center point of the trust region as follows. 

Let the reference set of diverse elites be $X_{div}$, then, $X_{\tau} \subset X_{tr}$ are the solutions in the trust region that satisfy the diversity criterion with $X_{div}$. The center of the trust region is updated to $x_{center}$ as below, 
\begin{equation}
    \label{eq:center-solution}
    x_{center} = \left\{
    \begin{array}{lcl}
        \arg_x \min_{x \in X_{\tau}} f(x) && X_{\tau}\neq \emptyset
        \\
        \arg_x \max_{x \in X_{tr},y \in X_{div}} div(x,y)
        && X_{\tau} = \emptyset
    \end{array} \right.
\end{equation}
%Given the X_{div} is the set of diverse elites from which the distance threshold should be maintained, and $X^_{\tau} \subseteq X_{tr}$ is the set of solutions in the trust region that maintains the distance from $X_{div}$, $x_{best}$ is the solution with the best objective value in $$
This definition of $x_{center}$ derives from Equation \ref{eq:best-div-solution}. This center selection allows the trust region to be adjusted considering both the quality and diversity. Thus, the new candidates are more likely to meet the diversity criteria. When none of the solutions in the trust region satisfy the minimum distance criterion ($X_{\tau}=\emptyset$, in consecutive iterations in the optimisation loop, it can be an indication that the current trust region is too close to previously selected diverse solutions in $X_{div}$. Therefore, if $X_{\tau}$ is found to be a null set three consecutive times at centre selection, divTuRBO1 triggers a restart of the trust region. Restarting trust region adopts the same steps as the standard TuRBO1 when its trust region reaches the minimum length thresholds (see Section \ref{sec:prelim}).

Secondly, divTuRBO1 also implements the diversity criteria when selecting the next batch of candidates from the predicted points. The acquisition function in TuRBO1 makes these selections according to the objective values predicted by the surrogate. In divTuRBO1, we make these selections considering both the objective values and whether the candidate maintains a minimum distance of $\tau$ with the reference set of solutions. If all candidate points violate the diversity requirement and are located closer to one or more reference points, we select the candidate that is located farthest. The changes in candidate selection allow the exploitation of the diverse candidates found by the surrogate model. The steps of the divTuRBO1 are given in the Algorithm \ref{alg:divTuRBO1}.

The goal of divTuRBO1 is to identify a single solution that maintains a given distance with reference solution points. To discover a desired number of $m$ solutions that are diverse from each other, we need to combine $m$ runs of the divTuRBO1 algorithm and incorporate the output of one algorithm run with another to define the input set of diverse elites. Moreover, we need to distribute the total evaluation budget over the $m$ runs of divTuRBO1. Despite the specific way of combining the run, during or after the execution, the best diverse solution is selected based on the objective function $f$ and the distance of the solutions to the currently identified best diverse solutions in each run. 
Let $X_{div}$ be the set of best diverse solutions that have been observed so far by $n \leq m$ individual divTuRBO1 runs, and $X$ be the set of solution vectors from which the best solution should be selected, we use Equation \ref{eq:best-div-solution} to choose the best diverse solution for a given $\tau$ value.

\subsection{Sequential algorithm}

We propose two ways to combine the divTuRBO1 runs to find $m$ diverse solutions for a given problem. We propose the first algorithm, divTURBO1-seq, which combines the $m$ runs of divTuRBO1 sequentially, such that each run depends on the outcome of the previous runs. The steps of this algorithm are presented in Algorithm \ref{alg:divTuRBO1-seq}. The first run of divTuRBO yields an empty set as the diverse elites, and this run is similar to the standard TuRBO1. At the end of each divTuRBO run, the best solution found (that satisfies the distance threshold) is added to the set of diverse elites $X_{div}$ and passed on to the consequent divTuRBO1 runs. At the end of $m$ divTuRBO runs, the diverse elites would be of size $m$, which represents the best set of solutions that maintains the minimum distance of $\tau$ with each other.

\subsection{Interleaving algorithm}

In our second algorithm, divTuRBO1-int, the $m$ divTuRBO1 runs are executed in an interleaving fashion. This algorithm is presented in Algorithm \ref{alg:divTuRBO1-int}. In this algorithm, each run of divTuRBO1 pauses after a certain number of evaluations and shares its current progress, which is the current diverse elite, with other runs. Then, it waits for the rest of the runs to share their elites again, resuming optimisation with a new trust region and the new set of diverse elites shared by other runs.

In divTURBO1-seq, the individual runs of divTuRBO1 start after their predecessor runs are complete, and the optimisation is diversified depending only on the predecessor results. In contrast, each divTuRBO1 run in divTuRBO1-int shares its interim progress with other divTuRBO1 runs, and the diversity criterion considers elites shared by all other ($m-1$) runs, not only the predecessor runs.

We also consider an existing BO approach for diversity ROBOT (Rank-Ordered Bayesian optimisation with trust regions) \cite{ROBOT-pmlr-v206-maus23a}. This is also a trust region-based approach that uses $m$ trust regions to optimise $m$ diverse solutions simultaneously. The regions are organised according to a ranking order as $T_0, \ldots, T_i, \ldots, T_m$. Each trust region $T_i$ is constrained by the higher-ranked regions $T_0, \ldots, T_{i-1}$. The centre of the trust region $T_i$ is chosen from the solutions in $T_i$, comparing their distance to the centres of the higher-ranked regions $T_0, \ldots, T_{i-1}$.

\begin{figure*}[!ht]
    \centering
    \tiny
    \begin{tabular}{c}
        {(a) divTuRBO1-seq} \\
        \resizebox{0.78\textwidth}{0.3\textwidth}{
        \includegraphics{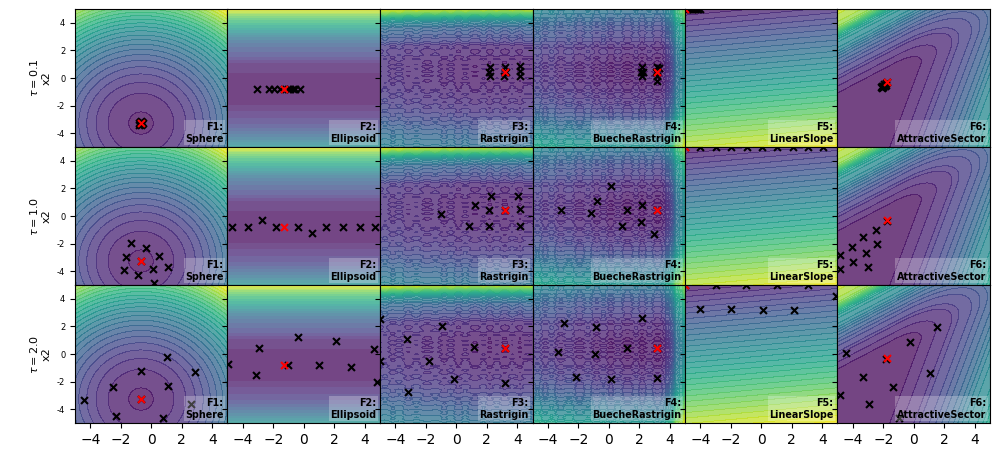}
        }
        \\
        {(b) divTuRBO1-int} \\
        \resizebox{0.78\textwidth}{0.3\textwidth}{
        \includegraphics{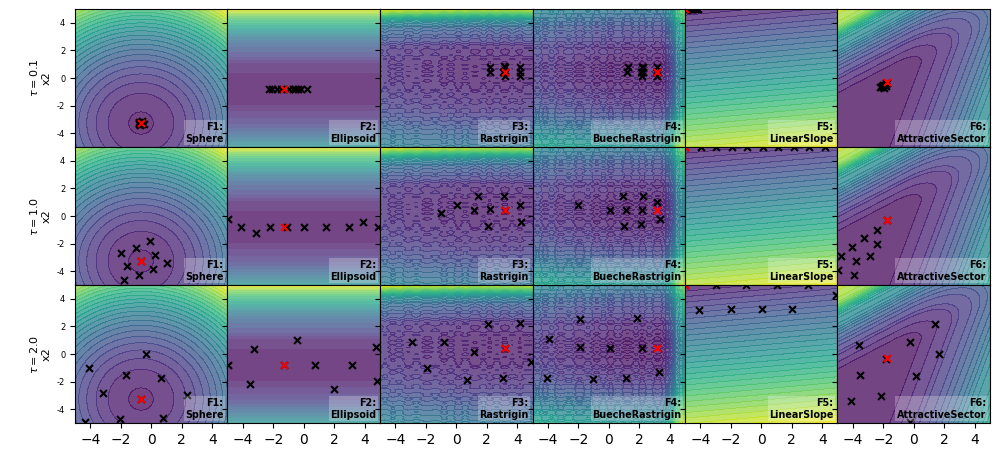}
        }
        \\
        {(c) ROBOT} \\
        \resizebox{0.78\textwidth}{0.3\textwidth}{
        \includegraphics{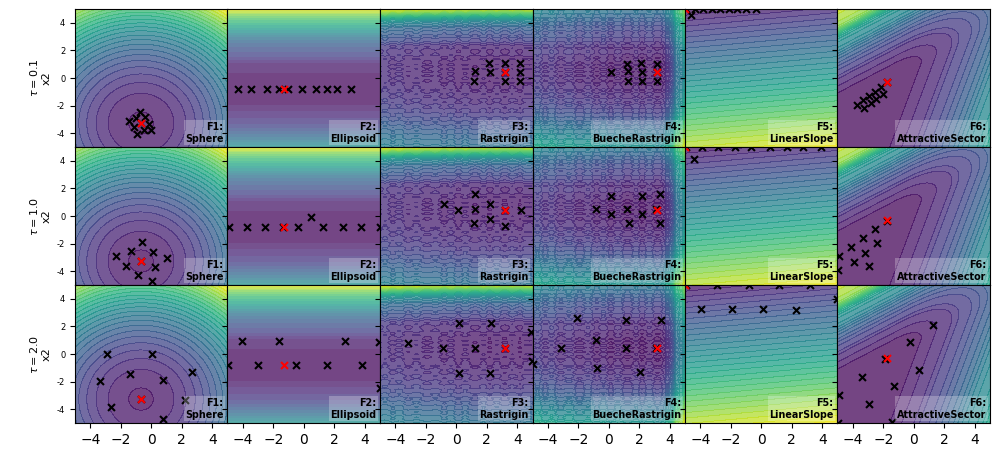}
        }  
         
    \end{tabular}
    \caption{Results from one run of each algorithm on the first 6 BBOB functions in the 2-D problem space for distance threshold values $\tau=0.1,1.0$ and $2.0$. The red and black crosses show the optimum of the functions and the outcome of the algorithms, respectively.}
    \label{fig:res_2d_mutli-tau}
    %\Description{Results from each algorithm for F1-6 in 2-D space for $\tau=0.1,1.0$ and $2.0$}
\end{figure*}

\begin{figure*}[!ht]
    \centering
    \tiny
    \begin{tabular}{c}
        {(a) divTuRBO1-seq} \\
        \resizebox{0.78\textwidth}{0.3\textwidth}{
        \includegraphics{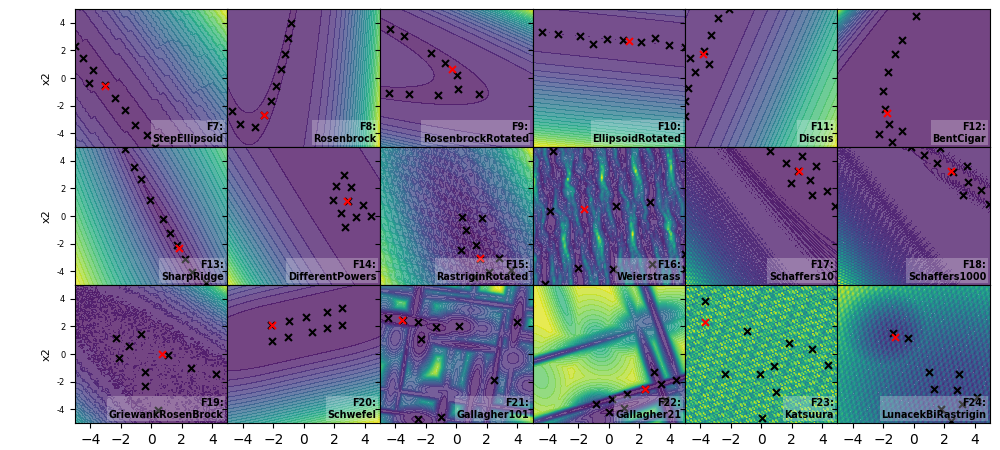}
        }
        \\
        {(b) divTuRBO1-int} \\
        \resizebox{0.78\textwidth}{0.3\textwidth}{
        \includegraphics{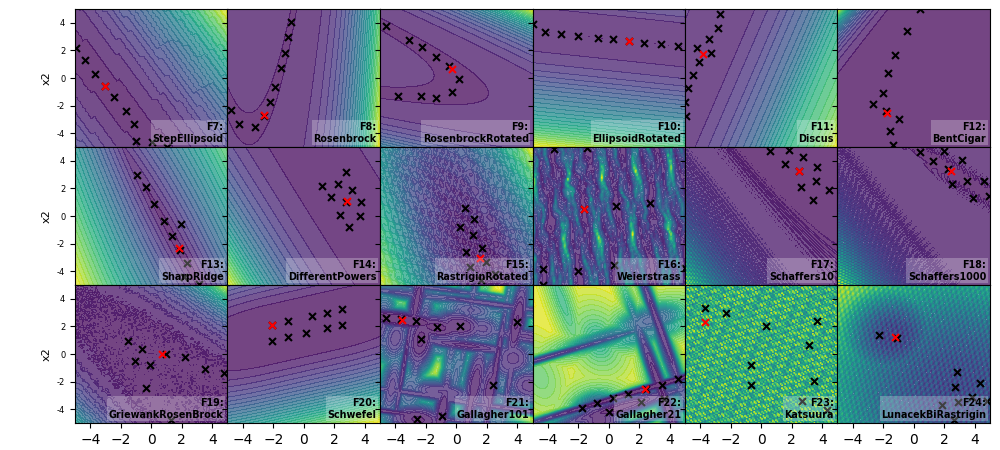}
        }
        \\
        {(c) ROBOT} \\
        \resizebox{0.78\textwidth}{0.3\textwidth}{
        \includegraphics{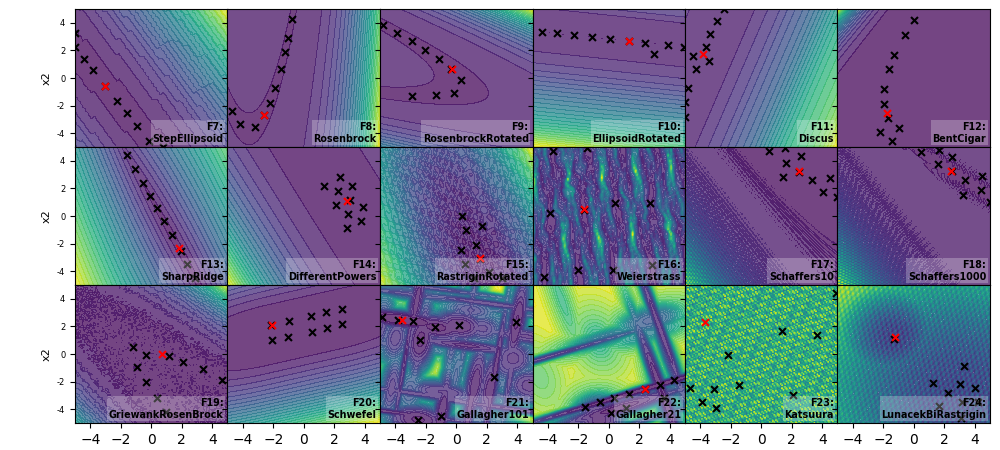}
        }
         
    \end{tabular}
    \caption{Results from one run of each algorithm on the 18 BBOB functions (F7-F24) for the 2-D problem space with distance threshold $\tau=1.0$. The red and black crosses show the optimum of the functions and the outcome of the algorithms, respectively.}
    \label{fig:res_2d_tau1}
    %\Description{Results from each algorithm on F7-F24 functions in the 2-D space with $\tau=1.0$.}
\end{figure*}

\section{Experiments\label{sec:experiments}}

We conduct our experiments on the 24 noiseless BBOB functions from COCO environment~\cite{coco-platform}. We consider these problems in dimensions $D = 2, 3, 10$ and $20$. The search space of these functions is a hypercube of $[-5,5]^D$. Most of these functions have the optimum in $[-4,4]^D$, but some exceptions exist, such as F5, where the optimum is on the domain boundary. Moreover, some of the functions have negative values for their optimum as  F1: -92.65, F7: -83.87, F8: -135.13, F9: -359.44, F10: -78.99, etc. For more details on the BBOB benchmark suite, we refer the reader to~\cite{coco-platform}. Our experiments aim to find $m=10$ diverse solutions using divTuRBO1-seq, divTuRBO1-int, and ROBOT algorithms on all 24 BBOB functions under selected dimensions. 

As we consider benchmark problems in various dimensions, ranging from 2 to 20, we determine the evaluation budget based on the number of dimensions. We allocate the evaluation budget for each algorithm run as follows. Let $D$ be the number of dimensions and $m$ be the number of diverse solutions; then the algorithm is allocated a total evaluation budget of  
$$(100 + 10\cdot D)\cdot m.$$ 

The multiple runs of divTuRBO1 within divTuRBO1-seq and divTuRBO1-int involve some computational overhead since each divTuRBO1 run maintains a separate surrogate model. However, these runs get only a portion of the total evaluation budget. Both divTuRBO1-seq and divTuRBO1-int  methods equally divide the allocated budget between the individual divTuRBO1 runs. Since we have $m=10$, individual divTuRBO1 gets one-tenth of the evaluation budget, which is $100+10\cdot D$. In our experiments, we consider the divTuRBO1-int to have five phases (unless otherwise specified). Therefore, each phase of divTuRBO1 inside divTuRBO1-int receives an equal budget divided among each phase as $20 + 2\cdot D$.

We conduct experiments considering different values for the minimum distance threshold. We mainly consider the diversity thresholds, $\tau$, as 0.1 and 1.0. Additionally, we conduct experiments for $\tau=2.0$ to illustrate how the results distribute in the search space for different $\tau$ values.

This comprehensive experimental evaluation considers 30 experiments from each algorithm for each setting. In each experiment, we record the mean objective value of the ten diverse solutions produced by each algorithm. The results are presented as a summary of these 30 mean values from the experiments. We also present the mean and standard deviation of the results from the 30 experiments. 
Moreover, we test for the statistical significance of the results from three algorithms. We use the Kruskal-Wallis test with 95\% confidence, followed by the Bonferroni correction. The statistical test results are presented in the tables in the following format. Let $X$ be the number assigned for each algorithm as in the column header; the statistical comparison $X^{+}$ or $X^{-}$ indicates that the method in the column outperforms $X$ or vice versa. If there is no significant difference between the two methods, the respective numbers do not appear.

\section{Results}\label{sec:results}

In this section, we analyse the results from the algorithms using visualisations of the results and statistical comparisons. 

\subsection{Visualisation of results for two-dimensional BBOB functions}
In this section, we run each selected algorithm on two-dimensional BBOB functions and visualise the results to analyse their behaviour for different $\tau$ values. Supplementary material presents a summary of these results, including statistical comparisons. 

Figure \ref{fig:res_2d_mutli-tau} shows results for 2-D functions: F1-F6 for distance thresholds $\tau=0.1, 1.0$ and $2.0$. The results in all the plots in Figure \ref{fig:res_2d_mutli-tau} comply with the diversity criterion and maintain the minimum distance $\tau$ between the ten solutions.
As the $\tau$ value is increased, the 10 solutions for each function spread over the search space, maintaining the given distance threshold. Although the results for higher $\tau$ values are similar across algorithms, we can see that the ROBOT results for $\tau=0.1$ behave differently from those of the other two algorithms. We can see this difference in behaviour when comparing the first row of plots (for $\tau=0.1$) in each subfigure in Figure \ref{fig:res_2d_mutli-tau}. While results from each algorithm maintain the desired level of diversity, ROBOT results for $\tau=0.1$ maintain a higher level of distance between the solutions compared to divTuRBO1-seq and divTuRBO1-int. This can affect the overall quality of the ROBOT results. 
\\
In Figure \ref{fig:res_2d_tau1}, we visualise the results for the remaining functions F7-F24 for $\tau=1.0$. These functions also maintain the diversity constraint, and the solutions lie in regions where the objective values are high, as indicated by the contour plots. We are interested in seeing if there are differences in the spread of the solutions from each algorithm. For most functions (F7-F22), the results look similar, and most of them have a solution overlapping with or lying close to the optimal solution of the function (marked as a red cross). The function F23 has a highly rugged search space, and the elite solutions are naturally located at some distance from each other. The results for this function from each algorithm spread very differently. Due to the rugged nature of the function, some solutions are placed at a higher distance than $\tau$ from other solutions. Also, none of the solutions from the three algorithms overlap with the optimum (marked in red). Compared to other functions, the solution closest to the optimum of F23 lies off the optimum. F24 is also an interesting function that maintains a multi-modal structure with two funnels. According to the results in Figure \ref{fig:res_2d_tau1}, due to the distance criterion, the 10 solutions are spread over both funnel areas. The two proposed algorithms, divTuRBO1-seq and divTuRBO1-seq, have found two solutions within the funnel area of the optimum and eight other solutions in the other funnel area. However, ROBOT has found only one solution near optimum, and the remaining nine solutions lie in the second funnel. Overall, all solutions maintain the distance threshold and are located where the objective values are reasonably high.

\begin{table*}[!t] \centering 
%\resizebox{0.5\textwidth}{!}%{0.6\textwidth}
\setlength{\columnsep}{-1pt}
\fontsize{6.2pt}{7.44pt}\selectfont 
{
\setlength{\columnsep}{0pt}
    \begin{tabular}{|l|r|r r r|r r r|r r r|} \hline
%& & \multicolumn{9}{|c|}{\kokila{Summary of 30 experiments on 20 dimensional BBOB functions}} \\\hline
~ & $ ~ $ & \multicolumn{3}{|c|}{divTuRBO1-seq (1)} & \multicolumn{3}{|c|}{divTuRBO1-int (2)} & \multicolumn{3}{|c|}{ROBOT (3)} \\ \cline{3-11}
 ~ & D & mean & st.dev & stat & mean & st.dev & stat & mean & st.dev & stat  \\ \hline

F-1 & 3 &	\textbf{-92.64} &	 0.00 &	 $  3^{+}$ &	 \textbf{-92.64} &	 0.00 &	 $  3^{+}$ &	 -92.54 &	 0.00 &	 $1^{-} 2^{-} $ \\	 
~ & 10 & 	\textbf{-92.62} &	 0.00 &	 $  3^{+}$ &	 \textbf{-92.62} &	 0.00 &	 $  3^{+}$ &	 -85.67 &	 0.00 &	 $1^{-} 2^{-} $ \\	 
~ & 20 & 	\textbf{-92.52} &	 0.01 &	 $  3^{+}$ &	 -92.50 &	 0.01 &	 $  3^{+}$ &	 -63.40 &	 0.00 &	 $1^{-} 2^{-} $ \\	\hline
F-2 & 3 &	\textbf{281.46} &	 1.12 &	 $  3^{+}$ &	 281.95 &	 1.31 &	 $  3^{+}$ &	 562.32 &	 0.00 &	 $1^{-} 2^{-} $ \\	 
~ & 10 & 	\textbf{2931.11} &	 588.87 &	 $  3^{+}$ &	 3103.10 &	 585.62 &	 $  3^{+}$ &	 30069.74 &	 0.00 &	 $1^{-} 2^{-} $ \\	 
~ & 20 & 	\textbf{11728.76} &	 1546.61 &	 $  3^{+}$ &	 13306.79 &	 2111.12 &	 $  3^{+}$ &	 351199.10 &	 0.00 &	 $1^{-} 2^{-} $ \\	\hline
F-3 & 3 &	 22.79 &	 0.21 &	 $  3^{+}$ &	\textbf{22.72} &	 0.13 &	 $  3^{+}$ &	 28.34 &	 0.00 &	 $1^{-} 2^{-} $ \\	 
~ & 10 &	\textbf{50.99} &	 3.04 &	 $  3^{+}$ &	 52.69 &	 1.76 &	 $  3^{+}$ &	 111.30 &	 0.00 &	 $1^{-} 2^{-} $ \\	 
~ & 20 & 	\textbf{127.22} &	 5.98 &	 $  3^{+}$ &	 134.74 &	 6.60 &	 $  3^{+}$ &	 290.85 &	 0.00 &	 $1^{-} 2^{-} $ \\	\hline
F-4 & 3 &	 25.66 &	 0.52 &	 $  3^{+}$ &	\textbf{25.47} &	 0.57 &	 $  3^{+}$ &	 31.63 &	 0.00 &	 $1^{-} 2^{-} $ \\	 
~ & 10 &	\textbf{75.00} &	 3.30 &	 $  3^{+}$ &	 77.51 &	 4.21 &	 $  3^{+}$ &	 143.70 &	 0.00 &	 $1^{-} 2^{-} $ \\	 
~ & 20 &	\textbf{180.21} &	 9.15 &	 $  3^{+}$ &	 190.02 &	 8.70 &	 $  3^{+}$ &	 404.47 &	 0.00 &	 $1^{-} 2^{-} $ \\	\hline
F-5 & 3 &	\textbf{51.96} &	 0.01 &	 $  3^{+}$ &	 51.98 &	 0.01 &	 $  3^{+}$ &	 53.47 &	 0.00 &	 $1^{-} 2^{-} $ \\	 
~ & 10 & 	\textbf{52.74} &	 0.06 &	 $  3^{+}$ &	 52.85 &	 0.12 &	 $  3^{+}$ &	 81.74 &	 0.00 &	 $1^{-} 2^{-} $ \\	 
~ & 20 & 	\textbf{57.15} &	 0.28 &	 $  3^{+}$ &	 57.64 &	 0.37 &	 $  3^{+}$ &	 171.11 &	 0.00 &	 $1^{-} 2^{-} $ \\	\hline
F-6 & 3 &	\textbf{83.73} &	 0.03 &	 $  3^{+}$ &	 83.77 &	 0.04 &	 $  3^{+}$ &	 85.13 &	 0.00 &	 $1^{-} 2^{-} $ \\	 
~ & 10 & 	\textbf{95.08} &	 2.93 &	 $  3^{+}$ &	 98.23 &	 2.00 &	 $  3^{+}$ &	 139.90 &	 0.00 &	 $1^{-} 2^{-} $ \\	 
~ & 20 & 	\textbf{93.30} &	 0.90 &	 $ 2^{+} 3^{+}$ &	 106.56 &	 1.88 &	 $1^{-}  3^{+}$ &	 15915.35 &	 0.00 &	 $1^{-} 2^{-} $ \\	\hline
F-7 & 3 &	\textbf{-83.84} &	 0.01 &	 $  3^{+}$ &	 \textbf{-83.84} &	 0.01 &	 $  3^{+}$ &	 -83.61 &	 0.00 &	 $1^{-} 2^{-} $ \\	 
~ & 10 & 	\textbf{-81.36} &	 0.38 &	 $  3^{+}$ &	 -81.29 &	 0.45 &	 $  3^{+}$ &	 -55.69 &	 0.00 &	 $1^{-} 2^{-} $ \\	 
~ & 20 & 	\textbf{-74.63} &	 1.06 &	 $  3^{+}$ &	 -73.02 &	 1.34 &	 $  3^{+}$ &	 44.31 &	 0.00 &	 $1^{-} 2^{-} $ \\	\hline
F-8 & 3 &	\textbf{-134.88} &	 0.05 &	 $  3^{+}$ &	 -134.84 &	 0.08 &	 $  3^{+}$ &	 -133.01 &	 0.00 &	 $1^{-} 2^{-} $ \\	 
~ & 10 & 	\textbf{-125.21} &	 1.06 &	 $  3^{+}$ &	 -123.96 &	 1.26 &	 $  3^{+}$ &	 1108.04 &	 0.00 &	 $1^{-} 2^{-} $ \\	 
~ & 20 & 	\textbf{-106.29} &	 3.97 &	 $ 2^{+} 3^{+}$ &	 -78.65 &	 6.05 &	 $1^{-}  3^{+}$ &	 10334.21 &	 0.00 &	 $1^{-} 2^{-} $ \\	\hline
F-9 & 3 &	 -359.06 &	 0.07 &	 $  3^{+}$ &	\textbf{-359.09} &	 0.12 &	 $  3^{+}$ &	 -355.53 &	 0.00 &	 $1^{-} 2^{-} $ \\	 
~ & 10 & 	 -342.00 &	 4.74 &	 $  3^{+}$ &	\textbf{-342.21} &	 3.54 &	 $  3^{+}$ &	 1051.09 &	 0.00 &	 $1^{-} 2^{-} $ \\	 
~ & 20 & 	\textbf{-301.91} &	 11.50 &	 $  3^{+}$ &	 -289.35 &	 5.51 &	 $  3^{+}$ &	 14812.77 &	 0.00 &	 $1^{-} 2^{-} $ \\	\hline
F-10 & 3 &	\textbf{-52.16} &	 4.76 &	 $  3^{+}$ &	 -46.45 &	 7.18 &	 $  3^{+}$ &	 38.45 &	 0.00 &	 $1^{-} 2^{-} $ \\	 
~ & 10 & 	\textbf{4383.08} &	 567.31 &	 $  3^{+}$ &	 4899.75 &	 574.76 &	 $  3^{+}$ &	 40835.85 &	 0.00 &	 $1^{-} 2^{-} $ \\	 
~ & 20 & 	\textbf{20779.81} &	 3515.17 &	 $  3^{+}$ &	 23282.76 &	 2227.49 &	 $  3^{+}$ &	 259548.73 &	 0.00 &	 $1^{-} 2^{-} $ \\	\hline
F-11 & 3 &	\textbf{-96.97} &	 1.56 &	 $  3^{+}$ &	 -96.69 &	 1.50 &	 $  3^{+}$ &	 -88.34 &	 0.00 &	 $1^{-} 2^{-} $ \\	 
~ & 10 & 	 -38.42 &	 5.11 &	 $  3^{-}$ &	 -39.87 &	 8.36 &	 $  3^{-}$ &	\textbf{-52.84} &	 0.00 &	 $1^{+} 2^{+} $ \\	 
~ & 20 & 	 31.02 &	 8.11 &	 $  3^{-}$ &	 27.91 &	 10.68 &	 ~  &	\textbf{23.74} &	 0.00 &	 $1^{+}  $ \\	\hline
F-12 & 3 &	\textbf{325.88} &	 5.90 &	 $  3^{+}$ &	 335.13 &	 13.94 &	 $  3^{+}$ &	 4871.85 &	 0.00 &	 $1^{-} 2^{-} $ \\	 
~ & 10 & 	\textbf{57227.99} &	 24752.46 &	 $  3^{+}$ &	 62231.30 &	 36390.60 &	 $  3^{+}$ &	 5645096.40 &	 0.00 &	 $1^{-} 2^{-} $ \\	 
~ & 20 & 	\textbf{287349.10} &	 130135.59 &	 $  3^{+}$ &	 591873.54 &	 108183.21 &	 $  3^{+}$ &	 29168573.00 &	 0.00 &	 $1^{-} 2^{-} $ \\	\hline
F-13 & 3 &	\textbf{-50.13} &	 0.18 &	 $  3^{+}$ &	 -49.96 &	 0.25 &	 $  3^{+}$ &	 -34.82 &	 0.00 &	 $1^{-} 2^{-} $ \\	 
~ & 10 & 	\textbf{-34.96} &	 0.96 &	 $  3^{+}$ &	 -33.32 &	 1.30 &	 $  3^{+}$ &	 387.86 &	 0.00 &	 $1^{-} 2^{-} $ \\	 
~ & 20 & 	\textbf{-13.48} &	 1.35 &	 $  3^{+}$ &	 -11.82 &	 2.08 &	 $  3^{+}$ &	 914.96 &	 0.00 &	 $1^{-} 2^{-} $ \\	\hline
F-14 & 3 &	 \textbf{-57.89} &	 0.00 &	 $  3^{+}$ &	\textbf{-57.89} &	 0.00 &	 $  3^{+}$ &	 -57.83 &	 0.00 &	 $1^{-} 2^{-} $ \\	 
~ & 10 & 	\textbf{-57.79} &	 0.02 &	 $  3^{+}$ &	 -57.77 &	 0.03 &	 $  3^{+}$ &	 -54.74 &	 0.00 &	 $1^{-} 2^{-} $ \\	 
~ & 20 & 	\textbf{-57.62} &	 0.05 &	 $  3^{+}$ &	 -57.56 &	 0.08 &	 $  3^{+}$ &	 -48.41 &	 0.00 &	 $1^{-} 2^{-} $ \\	\hline
F-15 & 3 &	\textbf{-42.84} &	 0.25 &	 $  3^{+}$ &	 -42.71 &	 0.28 &	 $  3^{+}$ &	 -36.29 &	 0.00 &	 $1^{-} 2^{-} $ \\	 
~ & 10 & 	\textbf{-13.97} &	 2.70 &	 $  3^{+}$ &	 -13.46 &	 2.48 &	 $  3^{+}$ &	 47.72 &	 0.00 &	 $1^{-} 2^{-} $ \\	 
~ & 20 & 	\textbf{52.86} &	 4.84 &	 $  3^{+}$ &	 59.96 &	 5.46 &	 $  3^{+}$ &	 255.33 &	 0.00 &	 $1^{-} 2^{-} $ \\	\hline
F-16 & 3 &	\textbf{-260.10} &	 0.03 &	 $  3^{+}$ &	 -260.09 &	 0.05 &	 $  3^{+}$ &	 -258.96 &	 0.00 &	 $1^{-} 2^{-} $ \\	 
~ & 10 & 	\textbf{-257.29} &	 0.31 &	 $  3^{+}$ &	 -257.15 &	 0.30 &	 $  3^{+}$ &	 -245.95 &	 0.00 &	 $1^{-} 2^{-} $ \\	 
~ & 20 & 	\textbf{-253.44} &	 0.45 &	 $  3^{+}$ &	 -252.78 &	 0.46 &	 $  3^{+}$ &	 -234.95 &	 0.00 &	 $1^{-} 2^{-} $ \\	\hline
F-17 & 3 &	\textbf{-38.55} &	 0.02 &	 $  3^{+}$ &	 -38.54 &	 0.02 &	 $  3^{+}$ &	 -37.99 &	 0.00 &	 $1^{-} 2^{-} $ \\	 
~ & 10 & 	\textbf{-37.62} &	 0.17 &	 $  3^{+}$ &	 -37.61 &	 0.20 &	 $  3^{+}$ &	 -35.62 &	 0.00 &	 $1^{-} 2^{-} $ \\	 
~ & 20 & 	\textbf{-36.60} &	 0.16 &	 $  3^{+}$ &	 -36.40 &	 0.24 &	 $  3^{+}$ &	 -33.79 &	 0.00 &	 $1^{-} 2^{-} $ \\	\hline
F-18 & 3 &	\textbf{-38.22} &	 0.06 &	 $  3^{+}$ &	 -38.17 &	 0.05 &	 $  3^{+}$ &	 -36.00 &	 0.00 &	 $1^{-} 2^{-} $ \\	 
~ & 10 & 	\textbf{-34.90} &	 0.22 &	 $  3^{+}$ &	 -34.62 &	 0.43 &	 $  3^{+}$ &	 -25.22 &	 0.00 &	 $1^{-} 2^{-} $ \\	 
~ & 20 & 	\textbf{-31.46} &	 0.87 &	 $  3^{+}$ &	 -30.46 &	 0.81 &	 $  3^{+}$ &	 -19.90 &	 0.00 &	 $1^{-} 2^{-} $ \\	\hline
F-19 & 3 &	\textbf{40.70} &	 0.03 &	 ~  &	 40.75 &	 0.08 &	 ~  &	 40.74 &	 0.00 &	 ~  \\	 
~ & 10 & 	 \textbf{44.30} &	 0.25 &	 $  3^{+}$ &	\textbf{44.30} &	 0.26 &	 $  3^{+}$ &	 47.05 &	 0.00 &	 $1^{-} 2^{-} $ \\	 
~ & 20 & 	\textbf{46.32} &	 0.13 &	 $  3^{+}$ &	 46.45 &	 0.25 &	 $  3^{+}$ &	 50.60 &	 0.00 &	 $1^{-} 2^{-} $ \\	\hline
F-20 & 3 &	\textbf{183.77} &	 0.06 &	 $  3^{+}$ &	 183.83 &	 0.07 &	 $  3^{+}$ &	 184.54 &	 0.00 &	 $1^{-} 2^{-} $ \\	 
~ & 10 & 	\textbf{184.97} &	 0.09 &	 $  3^{+}$ &	 185.04 &	 0.08 &	 $  3^{+}$ &	 290.40 &	 0.00 &	 $1^{-} 2^{-} $ \\	 
~ & 20 & 	\textbf{185.45} &	 0.06 &	 $  3^{+}$ &	 185.48 &	 0.06 &	 $  3^{+}$ &	 4021.55 &	 0.00 &	 $1^{-} 2^{-} $ \\	\hline
F-21 & 3 &	 310.70 &	 0.06 &	 ~  &	 310.68 &	 0.06 &	 ~  &	\textbf{310.64} &	 0.00 &	 ~  \\	 
~ & 10 & 	\textbf{311.33} &	 0.18 &	 $  3^{+}$ &	 311.35 &	 0.41 &	 $  3^{+}$ &	 322.15 &	 0.00 &	 $1^{-} 2^{-} $ \\	 
~ & 20 & 	\textbf{311.39} &	 0.12 &	 $  3^{+}$ &	 311.74 &	 0.18 &	 $  3^{+}$ &	 341.26 &	 0.00 &	 $1^{-} 2^{-} $ \\	\hline
F-22 & 3 &	 43.08 &	 0.05 &	 $  3^{-}$ &	 43.06 &	 0.06 &	 ~  &	\textbf{43.02} &	 0.00 &	 $1^{+}  $ \\	 
~ & 10 & 	\textbf{44.50} &	 0.33 &	 $  3^{+}$ &	 44.52 &	 0.21 &	 $  3^{+}$ &	 60.05 &	 0.00 &	 $1^{-} 2^{-} $ \\	 
~ & 20 & 	 45.36 &	 0.51 &	 $  3^{+}$ &	\textbf{45.26} &	 0.19 &	 $  3^{+}$ &	 83.83 &	 0.00 &	 $1^{-} 2^{-} $ \\	\hline
F-23 & 3 &	 211.84 &	 0.21 &	 $  3^{-}$ &	 211.83 &	 0.19 &	 $  3^{-}$ &	\textbf{211.70} &	 0.00 &	 $1^{+} 2^{+} $ \\	 
~ & 10 & 	 212.56 &	 0.12 &	 $  3^{-}$ &	 212.49 &	 0.15 &	 $  3^{-}$ &	\textbf{212.26} &	 0.00 &	 $1^{+} 2^{+} $ \\	 
~ & 20 & 	 213.53 &	 0.19 &	 $  3^{-}$ &	 213.50 &	 0.17 &	 $  3^{-}$ &	\textbf{213.03} &	 0.00 &	 $1^{+} 2^{+} $ \\	\hline
F-24 & 3 &	\textbf{53.19} &	 0.66 &	 ~  &	 53.25 &	 0.57 &	 ~  &	 53.78 &	 0.00 &	 ~  \\	 
~ & 10 & 	 109.70 &	 1.38 &	 $  3^{+}$ &	\textbf{108.53} &	 2.85 &	 $  3^{+}$ &	 149.19 &	 0.00 &	 $1^{-} 2^{-} $ \\	 
~ & 20 & 	\textbf{204.37} &	 3.45 &	 $  3^{+}$ &	 209.89 &	 3.66 &	 $  3^{+}$ &	 332.42 &	 0.00 &	 $1^{-} 2^{-} $ \\	\hline

\end{tabular}
}
\caption{Results for 3-D, 10-D and 20-D BBOB functions for $\tau$  =0.1} \label{tab:res_tau_0-1} 
\end{table*}
\begin{table*}[!t] \centering

\setlength{\columnsep}{-1pt}
\fontsize{6.2pt}{7.44pt}\selectfont 
%\fontsize{9.5pt}{10.25pt}\selectfont 
%\resizebox{0.7\textwidth}{0.6\textwidth}
{
\setlength{\columnsep}{0pt}
    \begin{tabular}{|l|r|r r r|r r r|r r r|} \hline
%& & \multicolumn{9}{|c|}{\kokila{Summary of 30 experiments on 20 dimensional BBOB functions}} \\\hline
~ & $ ~ $ & \multicolumn{3}{|c|}{divTuRBO1-seq (1)} & \multicolumn{3}{|c|}{divTuRBO1-int (2)} & \multicolumn{3}{|c|}{ROBOT (3)} \\ \cline{3-11}
 ~ & $D$ & mean & st.dev & stat & mean & st.dev & stat & mean & st.dev & stat  \\ \hline

F-1 & 3 &	 -91.72 &	 0.03 &	 ~  &	\textbf{-91.73} &	 0.02 &	 ~  &	 -91.71 &	 0.00 &	 ~  \\	 
~ & 10 & 	\textbf{-91.91} &	 0.03 &	 $  3^{+}$ &	 -91.90 &	 0.03 &	 $  3^{+}$ &	 -85.67 &	 0.00 &	 $1^{-} 2^{-} $ \\	 
~ & 20 & 	 -92.03 &	 0.04 &	 $  3^{+}$ &	\textbf{-92.04} &	 0.03 &	 $  3^{+}$ &	 -63.40 &	 0.00 &	 $1^{-} 2^{-} $ \\	\hline
F-2 & 3 &	 406.73 &	 40.15 &	 $  3^{+}$ &	\textbf{393.27} &	 30.22 &	 $  3^{+}$ &	 805.53 &	 0.00 &	 $1^{-} 2^{-} $ \\	 
~ & 10 & 	\textbf{2485.78} &	 368.78 &	 $  3^{+}$ &	 2800.64 &	 406.14 &	 $  3^{+}$ &	 30069.74 &	 0.00 &	 $1^{-} 2^{-} $ \\	 
~ & 20 & 	\textbf{11185.32} &	 1543.96 &	 $  3^{+}$ &	 12162.29 &	 1152.99 &	 $  3^{+}$ &	 351199.10 &	 0.00 &	 $1^{-} 2^{-} $ \\	\hline
F-3 & 3 &	\textbf{25.78} &	 0.24 &	 $  3^{+}$ &	 26.07 &	 0.54 &	 $  3^{+}$ &	 27.63 &	 0.00 &	 $1^{-} 2^{-} $ \\	 
~ & 10 &	\textbf{50.74} &	 1.41 &	 $  3^{+}$ &	 52.30 &	 1.96 &	 $  3^{+}$ &	 111.30 &	 0.00 &	 $1^{-} 2^{-} $ \\	 
~ & 20 & 	\textbf{129.23} &	 7.66 &	 $  3^{+}$ &	 139.13 &	 9.42 &	 $  3^{+}$ &	 290.85 &	 0.00 &	 $1^{-} 2^{-} $ \\	\hline
F-4 & 3 &	 31.59 &	 1.60 &	 $  3^{+}$ &	\textbf{31.49} &	 0.64 &	 $  3^{+}$ &	 33.38 &	 0.00 &	 $1^{-} 2^{-} $ \\	 
~ & 10 &	\textbf{73.73} &	 3.76 &	 $  3^{+}$ &	 77.41 &	 5.20 &	 $  3^{+}$ &	 143.70 &	 0.00 &	 $1^{-} 2^{-} $ \\	 
~ & 20 &	\textbf{186.78} &	 10.36 &	 $  3^{+}$ &	 193.30 &	 11.84 &	 $  3^{+}$ &	 404.47 &	 0.00 &	 $1^{-} 2^{-} $ \\	\hline
F-5 & 3 &	\textbf{55.03} &	 0.02 &	 $  3^{+}$ &	 55.07 &	 0.04 &	 $  3^{+}$ &	 56.30 &	 0.00 &	 $1^{-} 2^{-} $ \\	 
~ & 10 & 	\textbf{54.88} &	 0.14 &	 $  3^{+}$ &	 55.06 &	 0.15 &	 $  3^{+}$ &	 81.74 &	 0.00 &	 $1^{-} 2^{-} $ \\	 
~ & 20 & 	\textbf{57.84} &	 0.42 &	 $  3^{+}$ &	 58.34 &	 0.49 &	 $  3^{+}$ &	 171.11 &	 0.00 &	 $1^{-} 2^{-} $ \\	\hline
F-6 & 3 &	\textbf{89.88} &	 0.46 &	 $ 2^{+} $ &	 90.47 &	 0.46 &	 $1^{-}  $ &	 90.18 &	 0.00 &	 ~  \\	 
~ & 10 & 	\textbf{96.22} &	 1.55 &	 $  3^{+}$ &	 99.39 &	 1.76 &	 $  3^{+}$ &	 139.90 &	 0.00 &	 $1^{-} 2^{-} $ \\	 
~ & 20 & 	\textbf{93.60} &	 0.88 &	 $ 2^{+} 3^{+}$ &	 105.39 &	 1.81 &	 $1^{-}  3^{+}$ &	 15915.35 &	 0.00 &	 $1^{-} 2^{-} $ \\	\hline
F-7 & 3 &	\textbf{-82.47} &	 0.10 &	 $  3^{+}$ &	 -82.44 &	 0.16 &	 $  3^{+}$ &	 -82.25 &	 0.00 &	 $1^{-} 2^{-} $ \\	 
~ & 10 & 	\textbf{-81.37} &	 0.17 &	 $  3^{+}$ &	 -81.21 &	 0.36 &	 $  3^{+}$ &	 -55.69 &	 0.00 &	 $1^{-} 2^{-} $ \\	 
~ & 20 & 	\textbf{-73.91} &	 1.02 &	 $  3^{+}$ &	 -73.19 &	 1.63 &	 $  3^{+}$ &	 44.31 &	 0.00 &	 $1^{-} 2^{-} $ \\	\hline
F-8 & 3 &	 -117.08 &	 4.60 &	 $  3^{+}$ &	\textbf{-117.38} &	 3.70 &	 $  3^{+}$ &	 -112.97 &	 0.00 &	 $1^{-} 2^{-} $ \\	 
~ & 10 & 	\textbf{-105.81} &	 4.42 &	 $  3^{+}$ &	 -101.78 &	 4.70 &	 $  3^{+}$ &	 1108.04 &	 0.00 &	 $1^{-} 2^{-} $ \\	 
~ & 20 & 	\textbf{-82.12} &	 6.78 &	 $  3^{+}$ &	 -71.30 &	 5.18 &	 $  3^{+}$ &	 10334.21 &	 0.00 &	 $1^{-} 2^{-} $ \\	\hline
F-9 & 3 &	 -350.44 &	 6.56 &	 ~  &	 -352.47 &	 5.19 &	 ~  &	\textbf{-352.91} &	 0.00 &	 ~  \\	 
~ & 10 & 	\textbf{-318.54} &	 3.50 &	 $  3^{+}$ &	 -318.30 &	 5.41 &	 $  3^{+}$ &	 1051.09 &	 0.00 &	 $1^{-} 2^{-} $ \\	 
~ & 20 & 	\textbf{-293.32} &	 7.15 &	 $  3^{+}$ &	 -277.46 &	 10.87 &	 $  3^{+}$ &	 14812.77 &	 0.00 &	 $1^{-} 2^{-} $ \\	\hline
F-10 & 3 &	\textbf{68.63} &	 31.30 &	 $  3^{+}$ &	 106.31 &	 38.64 &	 $  3^{+}$ &	 185.21 &	 0.00 &	 $1^{-} 2^{-} $ \\	 
~ & 10 & 	\textbf{4592.34} &	 1064.50 &	 $  3^{+}$ &	 4614.17 &	 1139.65 &	 $  3^{+}$ &	 40835.85 &	 0.00 &	 $1^{-} 2^{-} $ \\	 
~ & 20 & 	\textbf{20655.05} &	 3178.62 &	 $  3^{+}$ &	 21170.21 &	 3433.81 &	 $  3^{+}$ &	 259548.73 &	 0.00 &	 $1^{-} 2^{-} $ \\	\hline
F-11 & 3 &	\textbf{-93.08} &	 1.68 &	 $  3^{+}$ &	 -92.14 &	 1.07 &	 $  3^{+}$ &	 -58.02 &	 0.00 &	 $1^{-} 2^{-} $ \\	 
~ & 10 & 	 -40.32 &	 4.86 &	 $  3^{-}$ &	 -41.85 &	 5.74 &	 $  3^{-}$ &	\textbf{-52.84} &	 0.00 &	 $1^{+} 2^{+} $ \\	 
~ & 20 & 	 31.63 &	 13.56 &	 ~  &	 29.31 &	 9.09 &	 ~  &	\textbf{23.74} &	 0.00 &	 ~  \\	\hline
F-12 & 3 &	\textbf{44765.40} &	 17425.86 &	 $  3^{+}$ &	 55651.73 &	 23228.20 &	 $  3^{+}$ &	 89210.35 &	 0.00 &	 $1^{-} 2^{-} $ \\	 
~ & 10 & 	\textbf{207557.34} &	 30013.14 &	 $  3^{+}$ &	 246366.96 &	 40613.90 &	 $  3^{+}$ &	 5645096.40 &	 0.00 &	 $1^{-} 2^{-} $ \\	 
~ & 20 & 	\textbf{362427.46} &	 78099.40 &	 $ 2^{+} 3^{+}$ &	 650133.68 &	 125103.63 &	 $1^{-}  3^{+}$ &	 29168573.00 &	 0.00 &	 $1^{-} 2^{-} $ \\	\hline
F-13 & 3 &	\textbf{-18.55} &	 1.25 &	 $ 2^{+} 3^{+}$ &	 -15.06 &	 2.31 &	 $1^{-}  $ &	 -16.10 &	 0.00 &	 $1^{-}  $ \\	 
~ & 10 & 	\textbf{-8.95} &	 4.30 &	 $  3^{+}$ &	 -5.11 &	 3.83 &	 $  3^{+}$ &	 387.86 &	 0.00 &	 $1^{-} 2^{-} $ \\	 
~ & 20 & 	\textbf{21.98} &	 2.47 &	 $  3^{+}$ &	 25.55 &	 4.30 &	 $  3^{+}$ &	 914.96 &	 0.00 &	 $1^{-} 2^{-} $ \\	\hline
F-14 & 3 &	 -57.21 &	 0.01 &	 ~  &	\textbf{-57.21} &	 0.01 &	 $  3^{+}$ &	 -57.19 &	 0.00 &	 $ 2^{-} $ \\	 
~ & 10 & 	\textbf{-57.60} &	 0.02 &	 $  3^{+}$ &	 -57.59 &	 0.02 &	 $  3^{+}$ &	 -54.74 &	 0.00 &	 $1^{-} 2^{-} $ \\	 
~ & 20 & 	\textbf{-57.59} &	 0.05 &	 $  3^{+}$ &	 -57.58 &	 0.06 &	 $  3^{+}$ &	 -48.41 &	 0.00 &	 $1^{-} 2^{-} $ \\	\hline
F-15 & 3 &	\textbf{-39.08} &	 0.49 &	 $  3^{+}$ &	 -38.82 &	 0.49 &	 $  3^{+}$ &	 -36.85 &	 0.00 &	 $1^{-} 2^{-} $ \\	 
~ & 10 & 	\textbf{-13.65} &	 2.55 &	 $  3^{+}$ &	 -12.23 &	 2.20 &	 $  3^{+}$ &	 47.72 &	 0.00 &	 $1^{-} 2^{-} $ \\	 
~ & 20 & 	\textbf{54.14} &	 3.54 &	 $  3^{+}$ &	 62.84 &	 6.42 &	 $  3^{+}$ &	 255.33 &	 0.00 &	 $1^{-} 2^{-} $ \\	\hline
F-16 & 3 &	\textbf{-260.06} &	 0.06 &	 $  3^{+}$ &	 -260.05 &	 0.05 &	 $  3^{+}$ &	 -258.88 &	 0.00 &	 $1^{-} 2^{-} $ \\	 
~ & 10 & 	\textbf{-257.21} &	 0.41 &	 $  3^{+}$ &	 -257.08 &	 0.40 &	 $  3^{+}$ &	 -245.95 &	 0.00 &	 $1^{-} 2^{-} $ \\	 
~ & 20 & 	\textbf{-253.47} &	 0.73 &	 $  3^{+}$ &	 -252.99 &	 0.62 &	 $  3^{+}$ &	 -234.95 &	 0.00 &	 $1^{-} 2^{-} $ \\	\hline
F-17 & 3 &	 -37.23 &	 0.31 &	 ~  &	 -37.45 &	 0.10 &	 ~  &	\textbf{-37.45} &	 0.00 &	 ~  \\	 
~ & 10 & 	\textbf{-37.64} &	 0.15 &	 $  3^{+}$ &	 -37.54 &	 0.17 &	 $  3^{+}$ &	 -35.62 &	 0.00 &	 $1^{-} 2^{-} $ \\	 
~ & 20 & 	\textbf{-36.61} &	 0.25 &	 $  3^{+}$ &	 -36.40 &	 0.29 &	 $  3^{+}$ &	 -33.79 &	 0.00 &	 $1^{-} 2^{-} $ \\	\hline
F-18 & 3 &	 -34.01 &	 0.84 &	 $  3^{-}$ &	 -35.33 &	 0.28 &	 $  3^{-}$ &	\textbf{-35.79} &	 0.00 &	 $1^{+} 2^{+} $ \\	 
~ & 10 & 	\textbf{-34.71} &	 0.45 &	 $  3^{+}$ &	 -34.60 &	 0.48 &	 $  3^{+}$ &	 -25.22 &	 0.00 &	 $1^{-} 2^{-} $ \\	 
~ & 20 & 	\textbf{-30.86} &	 1.00 &	 $  3^{+}$ &	 -30.04 &	 1.08 &	 $  3^{+}$ &	 -19.90 &	 0.00 &	 $1^{-} 2^{-} $ \\	\hline
F-19 & 3 &	 41.33 &	 0.38 &	 $  3^{-}$ &	 41.04 &	 0.12 &	 $  3^{-}$ &	\textbf{40.72} &	 0.00 &	 $1^{+} 2^{+} $ \\	 
~ & 10 & 	\textbf{44.25} &	 0.19 &	 $  3^{+}$ &	 44.36 &	 0.21 &	 $  3^{+}$ &	 47.05 &	 0.00 &	 $1^{-} 2^{-} $ \\	 
~ & 20 & 	\textbf{46.26} &	 0.25 &	 $  3^{+}$ &	 46.47 &	 0.19 &	 $  3^{+}$ &	 50.60 &	 0.00 &	 $1^{-} 2^{-} $ \\	\hline
F-20 & 3 &	\textbf{184.05} &	 0.08 &	 $ 2^{+} $ &	 184.14 &	 0.08 &	 $1^{-}  $ &	 184.12 &	 0.00 &	 ~  \\	 
~ & 10 & 	\textbf{185.00} &	 0.10 &	 $  3^{+}$ &	 185.02 &	 0.06 &	 $  3^{+}$ &	 290.40 &	 0.00 &	 $1^{-} 2^{-} $ \\	 
~ & 20 & 	\textbf{185.44} &	 0.08 &	 $  3^{+}$ &	 185.47 &	 0.09 &	 $  3^{+}$ &	 4021.55 &	 0.00 &	 $1^{-} 2^{-} $ \\	\hline
F-21 & 3 &	\textbf{311.58} &	 0.03 &	 $  3^{+}$ &	 311.65 &	 0.04 &	 $  3^{+}$ &	 311.85 &	 0.00 &	 $1^{-} 2^{-} $ \\	 
~ & 10 & 	\textbf{311.34} &	 0.26 &	 $  3^{+}$ &	 311.40 &	 0.30 &	 $  3^{+}$ &	 322.15 &	 0.00 &	 $1^{-} 2^{-} $ \\	 
~ & 20 & 	\textbf{311.63} &	 0.59 &	 $  3^{+}$ &	 311.70 &	 0.28 &	 $  3^{+}$ &	 341.26 &	 0.00 &	 $1^{-} 2^{-} $ \\	\hline
F-22 & 3 &	 43.66 &	 0.08 &	 ~  &	 43.71 &	 0.13 &	 ~  &	\textbf{43.62} &	 0.00 &	 ~  \\	 
~ & 10 & 	 44.93 &	 0.49 &	 $  3^{+}$ &	\textbf{44.76} &	 0.22 &	 $  3^{+}$ &	 60.05 &	 0.00 &	 $1^{-} 2^{-} $ \\	 
~ & 20 & 	\textbf{45.52} &	 0.24 &	 $  3^{+}$ &	 45.86 &	 0.66 &	 $  3^{+}$ &	 83.83 &	 0.00 &	 $1^{-} 2^{-} $ \\	\hline
F-23 & 3 &	 211.90 &	 0.09 &	 $  3^{-}$ &	 211.95 &	 0.12 &	 $  3^{-}$ &	\textbf{211.63} &	 0.00 &	 $1^{+} 2^{+} $ \\	 
~ & 10 & 	 212.58 &	 0.12 &	 $  3^{-}$ &	 212.66 &	 0.13 &	 $  3^{-}$ &	\textbf{212.26} &	 0.00 &	 $1^{+} 2^{+} $ \\	 
~ & 20 & 	 213.47 &	 0.12 &	 $  3^{-}$ &	 213.48 &	 0.13 &	 $  3^{-}$ &	\textbf{213.03} &	 0.00 &	 $1^{+} 2^{+} $ \\	\hline
F-24 & 3 &	 54.80 &	 0.77 &	 ~  &	\textbf{54.42} &	 0.53 &	 $  3^{+}$ &	 55.27 &	 0.00 &	 $ 2^{-} $ \\	 
~ & 10 & 	\textbf{109.16} &	 2.06 &	 $  3^{+}$ &	 109.93 &	 2.32 &	 $  3^{+}$ &	 149.19 &	 0.00 &	 $1^{-} 2^{-} $ \\	 
~ & 20 & 	\textbf{206.64} &	 4.46 &	 $  3^{+}$ &	 209.10 &	 4.31 &	 $  3^{+}$ &	 332.42 &	 0.00 &	 $1^{-} 2^{-} $ \\	\hline

\end{tabular}
}
\caption{Results for 3-D, 10-D and 20-D BBOB functions for $\tau$ =1.0} \label{tab:res_tau_1-0} 
\end{table*}

\subsection{Statistical comparison of results for 3, 10, and 20 dimensional BBOB functions}

In this subsection, we consider the results for dimensions 3 to 20, for which the visualisation of results is not possible, similar to 2-D problem instances. Here, we present a summary of the results from 30 runs of each algorithm for dimensions 3, 10, and 20, and compare the mean objective values of each diverse set using statistical tools.
We consider the problem instances under two different distance thresholds $\tau=0.1$ and $1.0$, and the corresponding results are presented in Table \ref{tab:res_tau_0-1} and \ref{tab:res_tau_1-0}, respectively.

Here, we use the mean of the objective value of the ten solutions to represent the quality of the diverse solution set from each algorithm run. Tables provide a summary of the 30 runs, showing the mean and standard deviation of the 30 mean values (displayed in column headers as 'mean' and 'st.dev'). The 'stat' columns show the statistical comparisons from the Kruskal-Wallis test (see Section \ref{sec:experiments}). We also recall divTuRBO1-int considers 5 phases when obtaining the results presented in Table \ref{tab:res_tau_0-1} and \ref{tab:res_tau_1-0}.

As Table \ref{tab:res_tau_0-1} shows, divTuRBO1-seq gives the best results for most settings. All three algorithms show similar results for a few functions with three dimensions. However, the results for dimensions 10 and 20 using the proposed algorithms show significant improvement over ROBOT, except for the multi-modal function F23, where ROBOT gives the best results.

Most functions with a low number of dimensions have better results from divTuRBO1-seq and divTuRBO1-int than from ROBOT when $\tau=1.0$ is considered. The statistical results in Table \ref{tab:res_tau_1-0} show that ROBOT produces similar results as the new algorithms for more 3-D functions when considering $\tau=1.0$ compared to $\tau=0.1$. According to the statistical tests, the new algorithms yield better results for 22 out of the 24 BBOB functions in both 10- and 20-dimensional spaces. The exceptions are the F11 and F23 functions, which yield better results from ROBOT. 

According to the above results, divTuRBO1-seq and divTuRBO1-int algorithms appear to produce similar results for both $\tau$ values considered in Table \ref{tab:res_tau_0-1} and \ref{tab:res_tau_1-0}. 

\subsection{Influence of the evaluation budget and number of optimisation phases.}%for budget comparison and phases comparison

\begin{figure*}[!t]
    \centering \small
    \begin{tabular}{c}
    
        \resizebox{0.8\textwidth}{0.45\textwidth}{\includegraphics{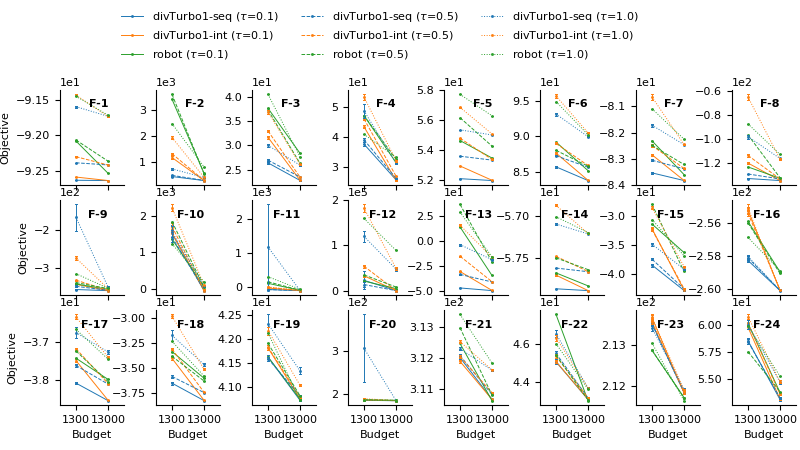}}
        \\
        (a) 3-D BBOB functions for $\tau = 0.1,0.5,$ and $1.0$	\\
        
        \resizebox{0.8\textwidth}{0.45\textwidth}{\includegraphics{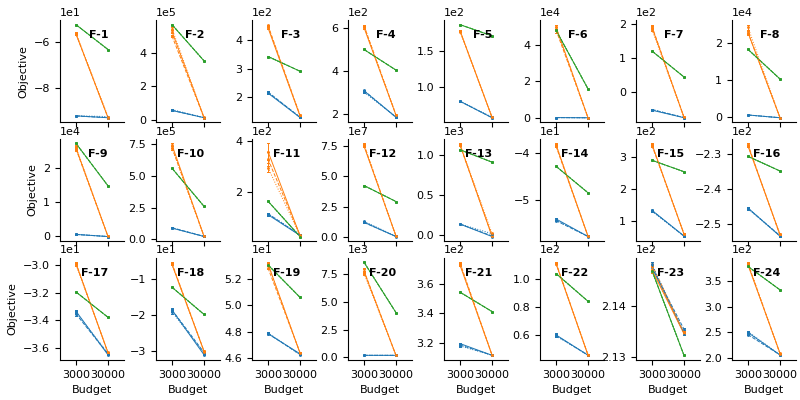}}
        \\
        (b) 20-D BBOB functions for $\tau = 0.1,0.5,$ and $1.0$	 \\
        
    \end{tabular}
    \caption{Results for BBOB functions using divTuRBO-seq, divTuRBO-int and ROBOT with different evaluation budgets.}
    %\Description{Results for BBOB functions using 3 algorithms with different evaluation budgets.}
    \label{fig:multi_budgets}
    
\end{figure*}

\begin{figure*}[!t]
    \centering
    \begin{tabular}{c}
        
        \resizebox{0.8\textwidth}{0.45\textwidth}{
        \includegraphics{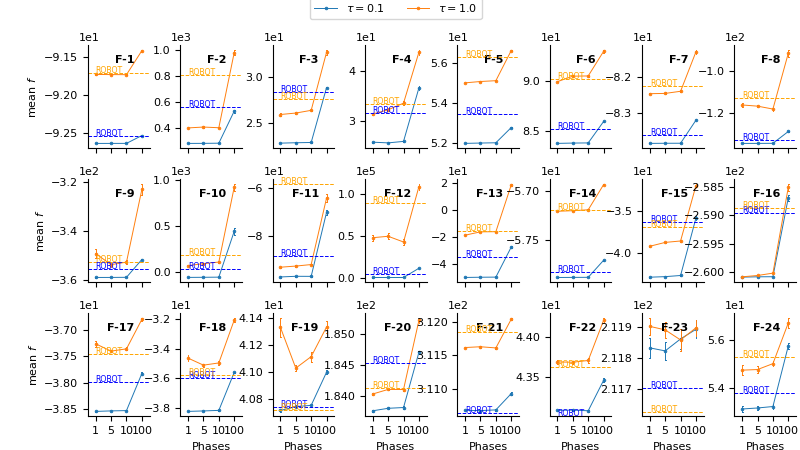}
        }
        \\
        {(a) 3 dimensional BBOB functions for $\tau = 0.1$ and $1.0$}
        \\
        \resizebox{0.8\textwidth}{0.45\textwidth}{
        \includegraphics{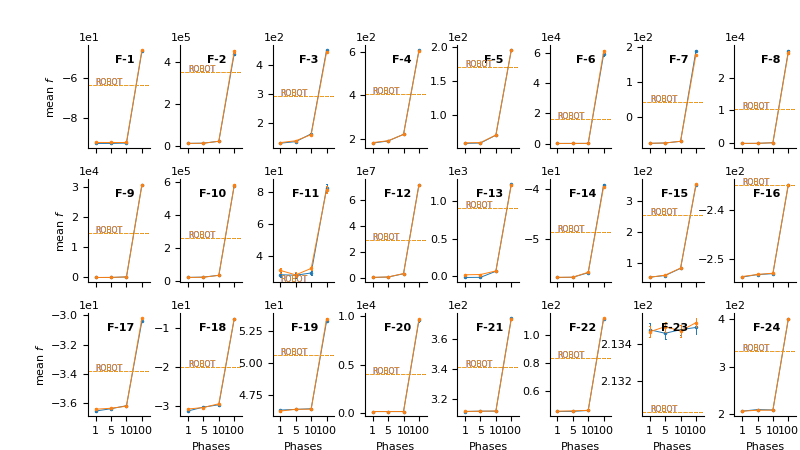}
        }
       \\
        {(b) 20 dimensional BBOB functions for $\tau = 0.1$ and $1.0$} \\
    \end{tabular}
    \caption{Results from divTuRBO-int when 1, 5, 10 and 100 phases for benchmark functions under $\tau=0.1$ and $1.0$. The dashed lines provide references to ROBOT results in each setting.}
    %\Description{Results from divTuRBO-int with different number of phases}
    \label{fig:multi_phase}
\end{figure*}

In general, BO methods require a small evaluation budget for optimisation. However, it is essential to consider the impact of the evaluation budget on the proposed algorithms. Selecting a subset of settings from the previous experiments, we investigate the impact of using a small budget and a sufficiently larger budget with the proposed algorithms and ROBOT. Figure \ref{fig:multi_budgets} presents the results for 3-D and 20-D problems using different evaluation budgets. 

In our experiments, we select the evaluation budget to be $(100 + 10\cdot D)\cdot m$ where $D$ and $m$ represent the number of dimensions and the size of the diverse solution set, respectively. For 3-D and 20-D benchmark problems, we use a total evaluation budget of 1300 and 3000 to obtain 10 diverse solutions. We compare the results of the algorithms when using this evaluation budget with the results when ten times this budget ($(1000 + 100\cdot D)\cdot m$) is used. The corresponding results are presented in Figure \ref{fig:multi_budgets}.

As Figure \ref{fig:multi_budgets} shows, the mean objective values have improved when using a higher budget. For 3-D benchmark functions, all algorithms have similar improvements in results. Most 3-D functions, such as F4, F8-12, F16-20, F23 and F24, show higher variation in results among the 30 runs with a lower evaluation budget. This variation drops when a larger budget is used, and 30 runs of the same algorithm generate similar results. The results on 3-D functions also demonstrate how the $\tau$ values affect the algorithm outcome. When the evaluation budget is low, the results are significantly different between different $\tau$ values for 3-D benchmark problems. For each 20-D function under different $\tau$ values, the results lie close together. The proposed methods outperform ROBOT for many functions when a larger budget is available. F11 gets similar results from all three algorithms with a higher budget. In contrast to all other functions, F23 gets the best results from ROBOT, with a higher number of evaluations showing significantly better than the proposed method. 

Next, we compare the influence of the number of phases in divTuRBO1-int on obtaining a diverse set of optimal solutions. When divTuRBO1-int is equivalent to divTuRBO1-seq when it uses only one phase. We consider the results of divTuRBO1-int for both single-phase (divTuRBO1-seq) and multi-phase settings (with 5, 10, and 100 phases) compared to ROBOT.
Figure \ref{fig:multi_phase} shows the results for these settings for 3-D and 20-D problem spaces. This figure shows that increasing the number of phases of divTuRBO1-int to a very large value, like 100, is not favourable. This is because these settings use a significantly low number of evaluations per phase. 

First, we consider the results for 3-D functions, as shown in Figure \ref{fig:multi_phase}. These results indicate that all algorithms yield poor solutions (those with higher values) as the minimum distance threshold ($\tau$) increases. ROBOT shows better results than the proposed methods for F9, F22 and F23 and for more functions when $\tau=0.1$. In comparison, results for 20-D settings show that the proposed approaches produce better results than ROBOT when using 1-5 phases for all settings except for functions F11 and F23.
F23 gets better results from ROBOT than from divTuRBO1-based approaches (similar to previous comparisons in Figure \ref{fig:multi_budgets}). As the number of dimensions is higher, most functions do not show a significant difference between their results using divTuRBO1-int with phases 1 and 5. Functions like F3-F5 and F11-F19 show that results are better when using 1 or 5 phases than when using 10 phases. Based on these results, we can strongly recommend both divTuRBO-seq and divTuRBO-int, with a small number of phases, for diversity optimisation over ROBOT.

\section{Conclusions}\label{sec:conclusion}

This study proposes diversity optimisation approaches based on the trust region-based BO methods introduced with TuRBO. TuRBO is the first algorithm to use trust regions with BO and shows significant efficiency in achieving great results with a small budget. Also, TuRBO performs well as the number of dimensions in the problem increases. We introduce two adaptations of TuRBO1 to consider diversity in BO and conduct experimental comparisons of the proposed methods with ROBOT, as a baseline BO method for diversity optimisation.

ROBOT performs well under a low distance threshold or for specific problems such as F23, where the elites are naturally distributed with some level of diversity, and the majority of the experimental settings favour the proposed algorithms. Also, ROBOT performs better on problems in a low number of dimensions (2 or 3). However, the two approaches proposed in the paper produce significantly better results in settings with more dimensions (20D settings). In some settings, it is hard to distinguish between results from the sequential and interleaving algorithms. Considering the overall results, we recommend the interleaving algorithm (divTuRBO-int) with a small number of phases, as well as the sequential algorithm (divTuRBO-seq), for diversity optimisation.

The experiments on a range of benchmark functions demonstrate the applicability of the proposed algorithms across a wide range of problems. This research shows that simple adaptation of trust region-based BO is very effective in addressing diversity optimisation. It would be beneficial to design efficient applications for real-world problems, such as those in molecular biology, engineering, and physics, where diversity is crucial.

\section*{Acknowledgements}
This work has been supported by the Australian Research Council through grant FT200100536 and the Adelaide-CNRS Research Mobility Scheme. The authors are grateful to Carola Doerr, Ishara Hewa Pathiranage, Elena Raponi and Maria Laura Santoni for initial discussions on the topic of this project.

\bibliographystyle{ACM-Reference-Format}
\bibliography{main.bib}

\newpage
\appendix
\section{Supplementary Results}

\begin{table*}[!h] \small \centering
    \begin{tabular}{|l|r|r r r|r r r|r r r|} \hline
%& & \multicolumn{9}{|c|}{\kokila{Summary of 30 experiments on 2 dimensional BBOB functions}} \\\hline
~ & $ ~ $ & \multicolumn{3}{|c|}{divTuRBO1-seq (1)} & \multicolumn{3}{|c|}{divTuRBO1-int (2)} & \multicolumn{3}{|c|}{ROBOT (3)} \\ \cline{3-11}
 Func & $\tau$ & mean & st.dev & stat & mean & st.dev & stat & mean & st.dev & stat  \\ \hline

F-1 & 0.1 &	\textbf{-92.63} &	 0.00 &	 $  3^{+}$ &	\textbf{-92.63} &	 0.00 &	 $  3^{+}$ &	 -92.62 &	 0.00 &	 $1^{-} 2^{-} $ \\	
~ & 1.0 &	 -91.05 &	 0.03 &	 $  3^{-}$ &	 -91.06 &	 0.02 &	 $  3^{-}$ &	\textbf{-91.09} &	 0.00 &	 $1^{+} 2^{+} $ \\	\hline
F-2 & 0.1 &	\textbf{277.50} &	 0.33 &	 ~  &	 277.60 &	 0.33 &	 ~  &	 277.61 &	 0.00 &	 ~  \\	
~ & 1.0 &	\textbf{48708.00} &	 22749.12 &	 ~  &	 58046.74 &	 19854.52 &	 ~  &	 49866.22 &	 0.00 &	 ~  \\	\hline
F-3 & 0.1 &	\textbf{22.42} &	 0.07 &	 $  3^{+}$ &	 22.47 &	 0.09 &	 $  3^{+}$ &	 22.86 &	 0.00 &	 $1^{-} 2^{-} $ \\	
~ & 1.0 &	 35.19 &	 1.92 &	 $  3^{-}$ &	 32.43 &	 2.02 &	 $  3^{-}$ &	\textbf{30.06} &	 0.00 &	 $1^{+} 2^{+} $ \\	\hline
F-4 & 0.1 &	\textbf{23.19} &	 0.15 &	 $  3^{+}$ &	 23.40 &	 0.14 &	 $  3^{+}$ &	 24.29 &	 0.00 &	 $1^{-} 2^{-} $ \\	
~ & 1.0 &	 40.54 &	 2.32 &	 $ 2^{-} 3^{-}$ &	 36.19 &	 1.28 &	 $1^{+}  3^{-}$ &	\textbf{32.43} &	 0.00 &	 $1^{+} 2^{+} $ \\	\hline
F-5 & 0.1 &	\textbf{52.03} &	 0.01 &	 $  3^{+}$ &	 52.05 &	 0.02 &	 $  3^{+}$ &	 52.21 &	 0.00 &	 $1^{-} 2^{-} $ \\	
~ & 1.0 &	\textbf{56.09} &	 0.01 &	 $ 2^{+} 3^{+}$ &	 56.13 &	 0.02 &	 $1^{-}  3^{+}$ &	 56.56 &	 0.00 &	 $1^{-} 2^{-} $ \\	\hline
F-6 & 0.1 &	\textbf{83.70} &	 0.02 &	 $  3^{+}$ &	\textbf{83.70} &	 0.03 &	 $  3^{+}$ &	 83.84 &	 0.00 &	 $1^{-} 2^{-} $ \\	
~ & 1.0 &	 94.70 &	 0.33 &	 ~  &	 95.12 &	 0.48 &	 $  3^{-}$ &	\textbf{94.47} &	 0.00 &	 $ 2^{+} $ \\	\hline
F-7 & 0.1 &	\textbf{-83.83} &	 0.01 &	 $  3^{+}$ &	\textbf{-83.83} &	 0.00 &	 $  3^{+}$ &	 -83.81 &	 0.00 &	 $1^{-} 2^{-} $ \\	
~ & 1.0 &	\textbf{-77.96} &	 1.70 &	 ~  &	 -77.89 &	 1.10 &	 ~  &	 -76.63 &	 0.00 &	 ~  \\	\hline
F-8 & 0.1 &	 -135.05 &	 0.02 &	 ~  &	\textbf{-135.06} &	 0.02 &	 ~  &	\textbf{-135.06} &	 0.00 &	 ~  \\	
~ & 1.0 &	 -133.17 &	 0.35 &	 ~  &	 -133.35 &	 0.16 &	 ~  &	\textbf{-133.39} &	 0.00 &	 ~  \\	\hline
F-9 & 0.1 &	\textbf{-359.38} &	 0.01 &	 $  3^{+}$ &	 -359.36 &	 0.01 &	 $  3^{+}$ &	 -359.25 &	 0.00 &	 $1^{-} 2^{-} $ \\	
~ & 1.0 &	 -322.55 &	 16.84 &	 $ 2^{-} 3^{-}$ &	 -349.52 &	 7.11 &	 $1^{+}  3^{-}$ &	\textbf{-357.35} &	 0.00 &	 $1^{+} 2^{+} $ \\	\hline
F-10 & 0.1 &	\textbf{-77.90} &	 0.26 &	 $  3^{+}$ &	 -77.82 &	 0.29 &	 $  3^{+}$ &	 -65.48 &	 0.00 &	 $1^{-} 2^{-} $ \\	
~ & 1.0 &	 49843.31 &	 17491.10 &	 ~  &	 55374.55 &	 16956.90 &	 ~  &	\textbf{44909.97} &	 0.00 &	 ~  \\	\hline
F-11 & 0.1 &	\textbf{-100.00} &	 0.22 &	 $  3^{+}$ &	 -99.91 &	 0.34 &	 $  3^{+}$ &	 -96.87 &	 0.00 &	 $1^{-} 2^{-} $ \\	
~ & 1.0 &	 134635.53 &	 13699.20 &	 ~  &	 121179.66 &	 20633.62 &	 ~  &	\textbf{117820.19} &	 0.00 &	 ~  \\	\hline
F-12 & 0.1 &	\textbf{296.37} &	 0.24 &	 $  3^{+}$ &	\textbf{296.37} &	 0.31 &	 $  3^{+}$ &	 310.88 &	 0.00 &	 $1^{-} 2^{-} $ \\	
~ & 1.0 &	 60011.57 &	 36559.01 &	 ~  &	\textbf{52314.94} &	 41398.71 &	 ~  &	 85633.27 &	 0.00 &	 ~  \\	\hline
F-13 & 0.1 &	\textbf{-51.16} &	 0.10 &	 $  3^{+}$ &	 -51.10 &	 0.07 &	 $  3^{+}$ &	 -49.38 &	 0.00 &	 $1^{-} 2^{-} $ \\	
~ & 1.0 &	 6.88 &	 10.89 &	 $  3^{-}$ &	 -0.58 &	 4.43 &	 $  3^{-}$ &	\textbf{-22.94} &	 0.00 &	 $1^{+} 2^{+} $ \\	\hline
F-14 & 0.1 &	\textbf{-57.86} &	 0.00 &	 $  3^{+}$ &	\textbf{-57.86} &	 0.00 &	 $  3^{+}$ &	\textbf{-57.86} &	 0.00 &	 $1^{-} 2^{-} $ \\	
~ & 1.0 &	\textbf{-56.76} &	 0.01 &	 $ 2^{+} $ &	 -56.73 &	 0.02 &	 $1^{-}  3^{-}$ &	\textbf{-56.76} &	 0.00 &	 $ 2^{+} $ \\	\hline
F-15 & 0.1 &	\textbf{-43.27} &	 0.06 &	 $  3^{+}$ &	 -43.23 &	 0.05 &	 $  3^{+}$ &	 -42.36 &	 0.00 &	 $1^{-} 2^{-} $ \\	
~ & 1.0 &	 -34.45 &	 1.97 &	 $  3^{-}$ &	 -34.42 &	 1.84 &	 $  3^{-}$ &	\textbf{-36.45} &	 0.00 &	 $1^{+} 2^{+} $ \\	\hline
F-16 & 0.1 &	\textbf{-260.19} &	 0.01 &	 $  3^{+}$ &	 -260.18 &	 0.01 &	 $  3^{+}$ &	 -259.86 &	 0.00 &	 $1^{-} 2^{-} $ \\	
~ & 1.0 &	 -259.13 &	 0.66 &	 $  3^{-}$ &	 -259.68 &	 0.28 &	 $  3^{-}$ &	\textbf{-259.98} &	 0.00 &	 $1^{+} 2^{+} $ \\	\hline
F-17 & 0.1 &	\textbf{-38.48} &	 0.01 &	 $  3^{+}$ &	\textbf{-38.48} &	 0.01 &	 $  3^{+}$ &	 -38.38 &	 0.00 &	 $1^{-} 2^{-} $ \\	
~ & 1.0 &	\textbf{-36.41} &	 0.06 &	 ~  &	 -36.39 &	 0.06 &	 ~  &	\textbf{-36.41} &	 0.00 &	 ~  \\	\hline
F-18 & 0.1 &	\textbf{-38.23} &	 0.05 &	 $  3^{+}$ &	 -38.19 &	 0.04 &	 $  3^{+}$ &	 -37.76 &	 0.00 &	 $1^{-} 2^{-} $ \\	
~ & 1.0 &	\textbf{-26.69} &	 0.68 &	 $ 2^{+} 3^{+}$ &	 -25.63 &	 1.24 &	 $1^{-}  $ &	 -25.79 &	 0.00 &	 $1^{-}  $ \\	\hline
F-19 & 0.1 &	\textbf{40.48} &	 0.00 &	 $  3^{+}$ &	\textbf{40.48} &	 0.01 &	 $  3^{+}$ &	 40.49 &	 0.00 &	 $1^{-} 2^{-} $ \\	
~ & 1.0 &	 41.13 &	 0.10 &	 $ 2^{-} 3^{-}$ &	 40.75 &	 0.06 &	 $1^{+}  3^{-}$ &	\textbf{40.52} &	 0.00 &	 $1^{+} 2^{+} $ \\	\hline
F-20 & 0.1 &	\textbf{183.81} &	 0.02 &	 $  3^{+}$ &	 183.82 &	 0.02 &	 $  3^{+}$ &	 183.88 &	 0.00 &	 $1^{-} 2^{-} $ \\	
~ & 1.0 &	 185.20 &	 2.53 &	 $ 2^{+} $ &	 186.27 &	 2.73 &	 $1^{-}  $ &	\textbf{184.47} &	 0.00 &	 ~  \\	\hline
F-21 & 0.1 &	\textbf{310.62} &	 0.00 &	 $  3^{+}$ &	\textbf{310.62} &	 0.00 &	 ~  &	 310.63 &	 0.00 &	 $1^{-}  $ \\	
~ & 1.0 &	\textbf{311.68} &	 0.00 &	 $  3^{+}$ &	 311.69 &	 0.02 &	 ~  &	 311.71 &	 0.00 &	 $1^{-}  $ \\	\hline
F-22 & 0.1 &	\textbf{42.98} &	 0.00 &	 $  3^{+}$ &	\textbf{42.98} &	 0.00 &	 $  3^{+}$ &	\textbf{42.98} &	 0.00 &	 $1^{-} 2^{-} $ \\	
~ & 1.0 &	 43.86 &	 0.09 &	 ~  &	 43.97 &	 0.12 &	 $  3^{-}$ &	\textbf{43.83} &	 0.00 &	 $ 2^{+} $ \\	\hline
F-23 & 0.1 &	 211.38 &	 0.09 &	 $  3^{-}$ &	 211.36 &	 0.12 &	 $  3^{-}$ &	\textbf{211.19} &	 0.00 &	 $1^{+} 2^{+} $ \\	
~ & 1.0 &	 211.56 &	 0.33 &	 $  3^{-}$ &	 211.51 &	 0.17 &	 $  3^{-}$ &	\textbf{211.28} &	 0.00 &	 $1^{+} 2^{+} $ \\	\hline
F-24 & 0.1 &	 49.25 &	 0.22 &	 $  3^{+}$ &	\textbf{49.24} &	 0.30 &	 $  3^{+}$ &	 49.77 &	 0.00 &	 $1^{-} 2^{-} $ \\	
~ & 1.0 &	 51.12 &	 0.21 &	 $  3^{-}$ &	 51.28 &	 0.25 &	 $  3^{-}$ &	\textbf{50.78} &	 0.00 &	 $1^{+} 2^{+} $ \\	\hline

\end{tabular}
    \caption{Summary of mean values of the 10 diverse solutions obtained from 30 on 2-dimensional BBOB functions.} \label{tab:res_dim2} \end{table*}

%\subsubsection{Results for 2-dimensional settings}
Table \ref{tab:res_dim2} presents the results for 2-D benchmark functions, considering 30 runs of each algorithm. The results are formatted in a similar manner to the tables in the main paper. The mean value of the 10 diverse solutions is considered the result of each run. We consider 30 runs of each algorithm, and the table gives the mean and standard deviation of results from 30 runs in columns `mean' and `st.dev' and the statistical comparisons thereof in column `stat'.

\begin{figure*}[!t]
    \centering \small
    \begin{tabular}{c}
        \resizebox{0.8\textwidth}{0.45\textwidth}{\includegraphics{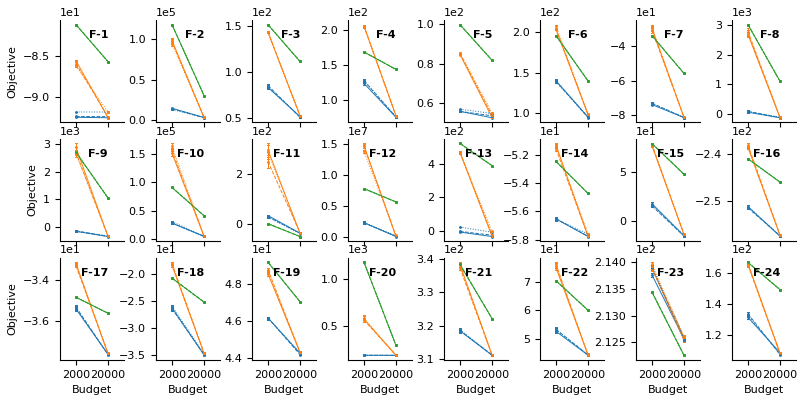}}
    \end{tabular}
    \caption{Results for 10-D functions using divTuRBO-seq, divTuRBO-int and ROBOT with different evaluation budgets.}
    \label{fig:appendix_budgets}
    
\end{figure*}

\begin{figure*}[!t]
    \centering
    \begin{tabular}{c}
        \resizebox{0.8\textwidth}{0.45\textwidth}{\includegraphics{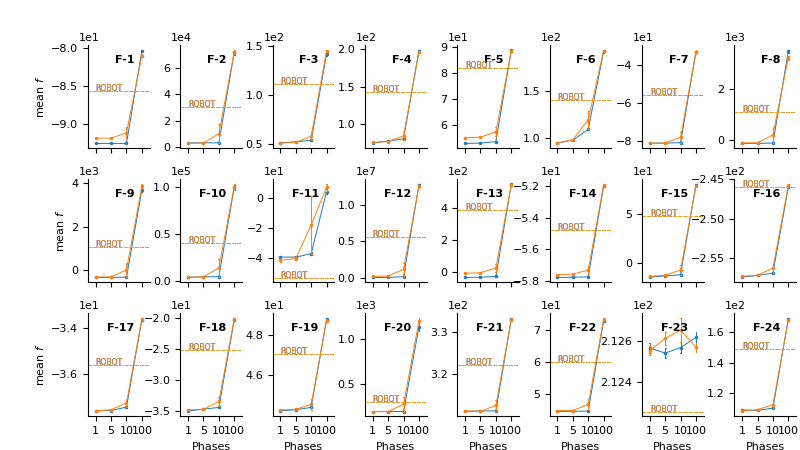}}
    \end{tabular}
    \caption{Results from divTuRBO-int when 1 (=divTuRBO1-seq), 5, 10 and 100 phases for 10-D benchmark functions under $\tau=0.1$ and $1.0$. The dashed lines provide references to ROBOT results in each setting.}
    \label{fig:appendix_phase}
\end{figure*}

Figure \ref{fig:appendix_budgets} corresponds Figure \ref{fig:multi_budgets} in the main paper. This figure shows results for benchmark functions in 10-D space when different budget sizes are used. 
Similarly, Figure \ref{fig:appendix_phase} relates to Figure \ref{fig:multi_phase} in the main paper and it shows results for benchmark functions in 10-D space when different number of phases are used in divTuRBO1-int.

\end{document}